\documentclass[twoside,11pt]{article}

\usepackage{jmlr2e} %JMLR

\usepackage{amsmath,bm,paralist,enumitem,booktabs,graphics,tikz,xcolor}
\usetikzlibrary{intersections}
\usepackage{srcltx}
\usepackage[mathscr]{eucal}
\usepackage[all]{xy}
\usepackage{tikz-cd}
\usetikzlibrary{arrows}
\usetikzlibrary{calc,decorations.pathmorphing,shapes}
\newcommand{\tb}{\textbf}   % bold text
\renewcommand{\b}{\mathbf}  % bold formula
\newcommand{\R}{\mathbb{R}}  % real numbers
\newcommand{\N}{\mathbb{N}}  % natural numbers = {1,2,...}
\renewcommand{\H}{\mathscr{H}} % RKHS
\renewcommand{\d}{\mathrm{d}} % dx
\renewcommand{\P}{\mathbb{P}} % probability measure-1
\newcommand{\Q}{\mathbb{Q}} % probability measure-2
\newcommand{\F}{\mathbb{F}} % finite signed measure
\newcommand{\X}{\mathscr{X}} %space-1
\newcommand{\Y}{\mathscr{Y}} %space-2
\newcommand{\Mb}{\mathscr{M}_b} %finite signed measure
\newcommand{\Mp}{\mathscr{M}_1^+} %probability measures
\newcommand{\B}{\mathscr{B}} %Borel
\newcommand{\Po}{\mathscr{P}} %power set
\newcommand{\I}{\mathcal{I}} %P_XY - P_X P_Y
\newcommand{\Ib}{\mathbb{I}} %I
\newcommand{\Fc}{\mathscr{F}} %F-characteristic
\newcommand{\vsubseteq}{\rotatebox{90}{\text{$\subseteq$}}} %vertical subseteq
\definecolor{gray1}{rgb}{0.97,0.97,0.97}
\definecolor{gray2}{rgb}{0.9,0.9,0.9}
\definecolor{gray3}{rgb}{0.75,0.75,0.75}
\definecolor{gray4}{rgb}{0.6,0.6,0.6}
\definecolor{gray5}{rgb}{0.45,0.45,0.45}

  \definecolor{OliveGreen}{rgb}{0.1,0.4,0.1}
\tikzset{
commutative diagrams/.cd,
arrow style=tikz,
diagrams={>=latex}
}

\usetikzlibrary{decorations.markings}
\tikzset{negated/.style={
        decoration={markings,
            mark= at position 0.85 with {
                \node[transform shape] (tempnode) {/};
            }
        },
        postaction={decorate}
    }
}

\allowdisplaybreaks %=> pagebreak is allowed for align

\makeatletter
 \renewcommand*{\@opargbegintheorem}[3]{\trivlist
  \item[\hskip \labelsep{\bfseries #1\ #2}] \textbf{(#3)}\ \itshape}
 \makeatother

\usepackage{lastpage}
\jmlrheading{18}{2018}{1-\pageref{LastPage}}{8/17; Revised 6/18}{7/18}{szabo18a}{Zolt{\'a}n Szab{\'o} and Bharath K. Sriperumbudur}
\ShortHeadings{Characteristic and Universal Tensor Product Kernels}{Szab{\'o} and Sriperumbudur}  %Short headings should be running head and authors last names
\firstpageno{1}

\begin{document}

\title{Characteristic and Universal Tensor Product Kernels}
\author{\name Zolt{\'a}n Szab{\'o} \email zoltan.szabo@polytechnique.edu\\
    \addr CMAP, {\'E}cole Polytechnique\\
    Route de Saclay, 91128 Palaiseau, France
    \AND
    \name Bharath K. Sriperumbudur \email bks18@psu.edu\\
    \addr Department of Statistics\\
    Pennsylvania State University\\
    314 Thomas Building\\
    University Park, PA 16802}

\editor{Francis Bach}

\maketitle
\begin{abstract}%  <- trailing '%' for backward compatibility of .sty file
Maximum mean discrepancy (MMD), also called energy distance or N-distance in statistics and Hilbert-Schmidt independence criterion (HSIC), specifically distance covariance in statistics, are among 
the most popular and successful approaches to quantify the difference and independence of random variables, respectively. 
Thanks to their kernel-based foundations, MMD and HSIC are applicable on a wide variety of domains. Despite their tremendous success, quite little is known about when 
HSIC characterizes independence and when MMD with tensor product kernel can discriminate probability distributions. 
In this paper, we answer these questions by studying various notions of characteristic property of the tensor product kernel. 
\end{abstract}


\section{Introduction}\label{Sec:intro}
Kernel methods \citep{scholkopf02learning} are among the most flexible and influential tools in machine learning and statistics, with superior performance demonstrated in a large number of areas and applications. The key idea in these methods is to map the data samples into a possibly infinite-dimensional feature space---precisely, a reproducing kernel Hilbert 
space (RKHS; \citealp{aronszajn50theory})---and apply linear methods in the feature space, without the explicit need to compute the map. A generalization of this idea to probability measures, i.e., mapping probability measures into an RKHS 
(\citealp[Chapter~4]{berlinet04reproducing}; \citealp{smola07hilbert}) has found novel applications in nonparametric statistics and machine learning. 
Formally, given a probability measure $\P$ defined on a measurable space $\X$ and an RKHS $\H_k$ with $k:\X \times \X \rightarrow \R$ as the reproducing kernel (which is symmetric and positive definite), $\P$ is embedded into $\H_k$ as
\begin{equation}
\P\mapsto\int_{\X} k(\cdot,x)\,\d\P(x)=:\mu_k(\P),\label{Eq:kernelmean}
\end{equation}
where $\mu_k(\P)$ is called the {\it{mean element}} or {\it{kernel mean embedding}} of $\P$. The \emph{mean embedding} of $\P$
has lead to a new generation of solutions in two-sample testing \citep{baringhaus04new,szekely04testing,szekely05new,borgwardt06integrating,harchaoui07testing,gretton12kernel}, 
goodness-of-fit testing \citep{chwialkowski16kernel,liu16kernelized,jitkrittum17linear,balasubramanian17optimality}, domain 
adaptation \citep{zhang13domain} and generalization \citep{blanchard17domain},
kernel belief propagation \citep{song11kernel},
kernel Bayes' rule \citep{fukumizu13kernel}, model criticism \citep{loyd14automatic,kim16examples},
approximate Bayesian computation \citep{park16k2abc},
probabilistic programming \citep{scholkopf15computing}, distribution classification \citep{muandet12learning,zaheer17deep}, 
distribution regression \citep{szabo16learning,law18bayesian} and topological data analysis \citep{kusano16persistence}. A recent survey on the topic is provided by \citet{maundet17kernel}.

Crucial to the success of the mean embedding based representation is whether it encodes all the information about the distribution, in other words whether the map in \eqref{Eq:kernelmean} is injective in which case the kernel is referred to as {\it{characteristic}}
\citep{fukumizu08kernel,sriperumbudur10hilbert}. Various characterizations for the characteristic property of $k$ is known in the literature \citep{fukumizu08kernel,fukumizu09kernel,sriperumbudur10hilbert,gretton12kernel} using which the popular kernels on $\R^d$ such as Gaussian, Laplacian, B-spline, inverse multiquadrics, and the Mat{\'e}rn class are shown to be characteristic. The characteristic property is closely related to the notion of {\it{universality}} (\citealp{steinwart01influence}; \citealp{micchelli06universal}; \citealp{carmeli10vector}; \citealp{sriperumbudur11universality})---$k$ is said to be universal if the corresponding RKHS $\H_k$ is dense in a certain target function class, for example, the class of continuous functions on compact domains---and the relation between these notions has recently 
been explored by \citet{sriperumbudur11universality,simon-gabriel16kernel}.

Based on the mean embedding in \eqref{Eq:kernelmean}, \citet{smola07hilbert} and \citet{gretton12kernel} defined a semi-metric, called the maximum mean discrepancy (MMD) on the space of probability measures: $$\text{MMD}_k(\P,\Q):=\Vert \mu_k(\P)-\mu_k(\Q)\Vert_{\H_k},$$
which is a metric iff $k$ is characteristic. A fundamental application of MMD is in non-parametric hypothesis testing that includes two-sample \citep{gretton12kernel} and independence tests \citep{gretton08kernel}. Particularly in independence testing, as a measure of independence, MMD measures the distance between the joint distribution $\P_{XY}$ and the product of marginals $\P_X\otimes\P_Y$ of two random variables $X$ and $Y$ which are respectively defined on measurable spaces $\X$ and $\Y$, with the kernel $k$ being defined on $\X\times\Y$. As aforementioned, if $k$ is characteristic, then $\text{MMD}_k(\P_{XY},\P_X\otimes\P_Y)=0$ implies $\P_{XY}=\P_X\otimes\P_Y$, i.e., $X$ and $Y$ are independent. A simple way to define a kernel on $\X\times\Y$ is through the tensor product of kernels $k_X$ and $k_Y$ defined on $\X$ and $\Y$ respectively: $k=k_X\otimes k_Y$, i.e., $k\left(\left(x,y\right),\left(x',y'\right)\right)=k_X(x,x')k_Y(y,y'),\,x,x'\in\X,\,y,y'\in\Y$, with the corresponding RKHS $\H_k=\H_{k_X}\otimes\H_{k_Y}$ being the tensor product space generated by $\H_{k_X}$ and $\H_{k_Y}$. This means, when $k=k_X\otimes k_Y$,
\begin{equation}\text{MMD}_k(\P_{XY},\P_X\otimes\P_Y)=\left\Vert \mu_{k_X\otimes k_Y}(\P_{XY})-\mu_{k_X\otimes k_Y}(\P_{X}\otimes\P_Y)\right\Vert_{\H_{k_X}\otimes\H_{k_Y}}.\label{Eq:MMD}
\end{equation}
In addition to the simplicity of defining a joint kernel $k$ on $\X\times\Y$, the tensor product kernel offers a principled way of combining
inner products ($k_X$ and $k_Y$) on domains that can correspond to different modalities (say images, texts, audio). By exploiting the isomorphism between tensor product Hilbert spaces and the space of 
Hilbert-Schmidt operators\footnote{In the equivalence one assumes that $\H_{k_X}$, $\H_{k_Y}$ are separable; this holds under mild conditions, for example if $\X$ and $\Y$ are separable topological domains and $k_X$, $k_Y$ are continuous 
\cite[Lemma~4.33]{steinwart08support}.}, it follows from \eqref{Eq:MMD} that
\begin{align}
  \text{MMD}_k(\P_{XY},\P_X\otimes\P_Y)=\Vert C_{XY}\Vert_{\text{HS}}=:\text{HSIC}_k(\P_{XY}), \label{eq:def:HSIC:M=2}
\end{align}
which is the Hilbert-Schmidt norm of the cross-covariance operator $C_{XY}:=\mu_{k_X\otimes k_Y}(\P_{XY})-\mu_{k_X}(\P_{X})\otimes\mu_{k_Y}(\P_Y)$
and is known as the {\it{Hilbert-Schmidt independence criterion}} (HSIC) \citep{gretton05measuring}. HSIC has enjoyed tremendous success in a variety of applications such
as independent component analysis \citep{gretton05measuring}, feature selection \citep{song12feature}, independence testing \citep{gretton08kernel,jitkrittum17adaptive2}, post selection inference \citep{yamada18post} and 
causal detection \citep{mooij16distinguishing,pfister17kernel,strobl17approximate}. Recently, MMD and HSIC (as defined in \eqref{eq:def:HSIC:M=2} for two components) have been shown by \citet{sejdinovic13equivalence} 
to be equivalent to other popular statistical measures such as the energy distance \citep{baringhaus04new,szekely04testing,szekely05new}---also known as N-distance \citep{zinger92characterization,klebanov05n-distance}---and distance covariance \citep{szekely07measuring,szekely09brownian,lyons13distance} respectively. HSIC has been generalized to $M\ge$ 2 components \citep{quadrianto09kernelized,sejdinovic13kernel} to measure the joint independence of $M$ random variables 
\begin{align}
\text{HSIC}_k\left(\P\right)=\left\Vert \mu_{\otimes^M_{m=1}k_m}(\P)-\otimes^M_{m=1}\mu_{k_m}\left(\P_m\right)\right\Vert_{\otimes^M_{m=1}\H_{k_m}},  \label{eq:def:HSIC:M>=2}
\end{align}
where $\P$ is a joint measure on the product space $\X:=\times^M_{m=1}\X_m$ and $\left(\P_m\right)^M_{m=1}$ are the marginal measures of $\P$ defined on $(\X_m)^M_{m=1}$ respectively.
The extended HSIC measure has recently been analyzed in the context of independence testing \citep{pfister17kernel}. In addition to testing, the extended HSIC measure is also useful in the problem of independent subspace analysis 
(ISA; \citealp{cardoso98multidimensional}), wherein the latent sources are separated by maximizing the 
degree of independence among them.
In all the applications of HSIC, the key requirement is that
 $k=\otimes^M_{m=1}k_m$ captures the joint independence of $M$ random variables (with joint distribution $\P$)---we call this property as $\I$-characteristic---, which is guaranteed if $k$ is characteristic. Since $k$ is defined in terms of $(k_m)^M_{m=1}$, it is of fundamental importance to understand the characteristic and $\I$-characteristic properties of $k$ in terms of the characteristic property of $(k_m)^M_{m=1}$, which is 
one of the main goals of this work.

For $M=2$, the characterization of independence, i.e., the $\I$-characteristic property of $k$, is studied by \citet{blanchard11generalizing} and \citet{gretton15simpler} where it has been 
shown that if $k_1$ and $k_2$ are universal, then $k$ is universal\footnote{\citet{blanchard11generalizing} deal with $c$-universal kernels while \citet{gretton15simpler} deals with $c_0$-universal kernels. 
A brief description of these notions are given in Section~\ref{sec:problem}. \citet{carmeli10vector, sriperumbudur10hilbert} provide further details on these notions of universality.} and therefore HSIC captures independence. 
A stronger version of this result can be obtained by combining \citep[Theorem 3.11]{lyons13distance} and \citep[Proposition 29]{sejdinovic13equivalence}: if $k_1$ and $k_2$ are characteristic, then the HSIC associated with $k=k_1\otimes k_2$ characterizes independence. Apart from these results, not much is known about the characteristic/$\I$-characteristic/universality properties of $k$ in terms of the individual kernels. 
Our \tb{goal} is to resolve this question and understand the characteristic, $\I$-characteristic and universal property of the product kernel ($\otimes_{m=1}^M k_m$) in terms of the kernel components ($(k_m)_{m=1}^M$) for $M\ge 2$. Because of the relatedness of MMD and HSIC to energy distance and distance covariance, our results also contribute to the better understanding of these other measures that are popular in the statistical literature.

Specifically, our results shed light on the following \tb{surprising phenomena} of the $\I$-characteristic property of $\otimes_{m=1}^M k_m$ for $M\ge 3$: 
\begin{compactenum}
    \item characteristic property of $(k_m)_{m=1}^M$ is not sufficient but necessary for $\otimes_{m=1}^M k_m$ to be $\I$-characteristic;
    \item universality of $(k_m)_{m=1}^M$ is sufficient for $\otimes_{m=1}^M k_m$ to be $\I$-characteristic, and
    \item if at least one of $(k_m)^M_{m=1}$ is only characteristic and not universal, then $\otimes_{m=1}^M k_m$ need not be $\I$-characteristic.
\end{compactenum}

\begin{figure}[t]
\label{fig:summary}
\begin{center}
\begin{tikzcd}[scale=1,row sep=1in,column sep=huge]
\otimes_0\text{-char}  \arrow[Leftarrow,blue]{r}{\eqref{eq:char-relations}} \arrow[Leftrightarrow,sloped,below,rotate=180,blue]{dd}[description]{\text{Remark}~\ref{Rem:remark}(iii)}
& \otimes\text{-char}  \arrow[Leftarrow,blue]{r}{\eqref{eq:char-relations}} \arrow[Rightarrow,sloped,xshift=0ex,yshift=.15ex,blue,bend left=20]{rr}[pos=0.5]{\text{Remark}~\ref{Rem:new}}
& \text{char} \arrow[Leftarrow,blue]{r}{\eqref{eq:char-relations}} \arrow[Rightarrow,xshift=.7ex,blue]{d}{\eqref{eq:char-relations}} \arrow[negated,Leftarrow,xshift=-.9ex,swap,red]{d}[yshift=-2ex]{\text{Example}~\ref{example1:k_m:char=does not=>k_1xk_2:char-or-x-char,but-here-I-char}}
& c_0\text{-universal}\arrow[Leftrightarrow,sloped,above,blue]{dd}[description]{\text{Theorem}~\ref{thm4:prod-of-c0universal}}\\
& & \I\text{-char}  & \\
(k_m)^M_{m=1}\,\,\text{char} \arrow[Rightarrow,negated,sloped,red]{ruu}[pos=0.7]{\text{Example}~\ref{example1:k_m:char=does not=>k_1xk_2:char-or-x-char,but-here-I-char}} \arrow[Rightarrow,negated,sloped,xshift=-1ex,red]{rruu}[pos=0.7]{\text{Example}~\ref{example1:k_m:char=does not=>k_1xk_2:char-or-x-char,but-here-I-char}} 
\arrow[Rightarrow,sloped,xshift=0ex,yshift=.15ex,blue,bend left=7]{rru}[pos=0.8]{\text{Theorem}~\ref{thm2:k_m:char<=>xk_m:I-char}\,\,(M=2)} \arrow[Rightarrow,sloped,negated,xshift=0ex,yshift=-.3ex,red]{rru}[description]{\text{Example}~\ref{example:k_m:char=does not=>x_m k_m:I-char}\,\,(M\ge 3)} \arrow[Rightarrow,negated,yshift=.7ex,red]{rrr}{\text{\citet{sriperumbudur11universality}}} \arrow[Leftarrow,sloped,xshift=-1ex,yshift=-.8ex,blue,bend right=7]{rru}[swap]{\text{Theorem}~\ref{thm2:k_m:char<=>xk_m:I-char}}\arrow[Leftarrow,yshift=-.7ex,blue]{rrr}[below]{\text{\citet{sriperumbudur11universality}}}
&&& (k_m)^M_{m=1}\,\,c_0 \text{-universal}
\end{tikzcd}
\end{center}\vspace{-1mm}
\caption{Summary of results: ``char" denotes characteristic. In addition to the usual characteristic property, three new notions $\otimes_0$-characteristic, $\otimes$-characteristic and $\I$-characteristic are introduced in Definition~\ref{def:F-char} which along with $c_0$-universal (in the top right corner) correspond to the property of the tensor product kernel $\otimes^M_{m=1}k_m$, while the bottom part of the picture corresponds to the individual kernels $(k_m)^M_{m=1}$ being characteristic or $c_0$-universal. If $(k_m)^M_{m=1}$-s are continuous, bounded and translation invariant kernels on $\R^{d_m},\,m\in[M]$, all the notions are equivalent (see Theorem~\ref{thm4:contboundedshiftinv}).}\vspace{-8mm}
\end{figure}
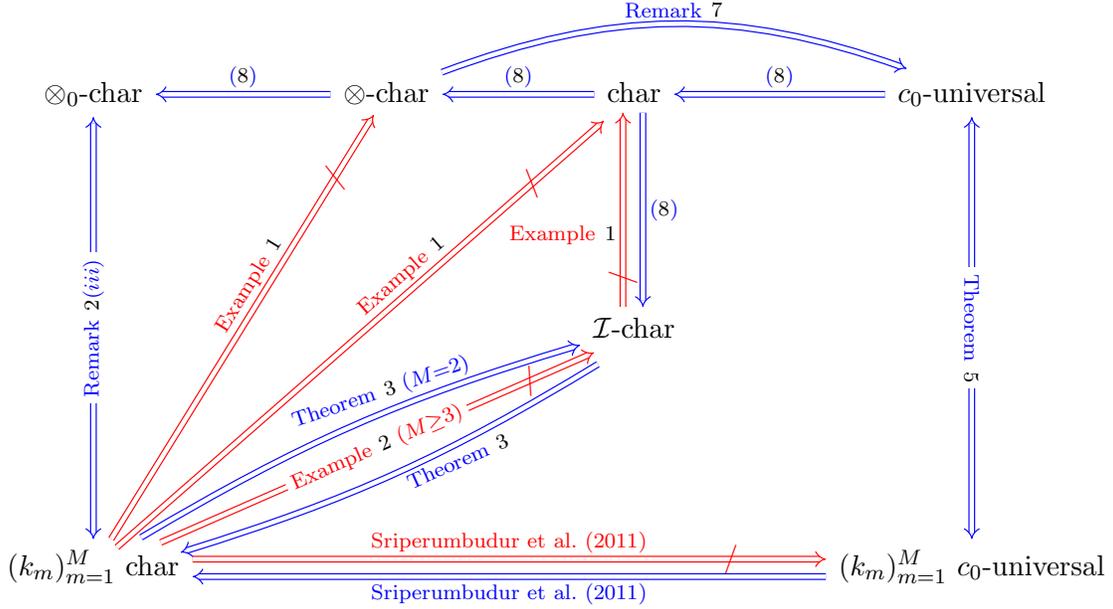

The paper is organized as follows. In Section~\ref{sec:problem}, we conduct a comprehensive analysis about the above mentioned properties of $k$ and $(k_m)^M_{m=1}$ for any positive integer $M$. To this end, we define various notions of characteristic property on the product space $\X$ (see Definition~\ref{def:F-char} and Figure~\ref{fig:F-char-demo}(a) in Section~\ref{sec:problem}) and explore the relation between them. In order to keep our presentation in this section to be non-technical, we relegate the problem formulation to Section~\ref{sec:problem}, with the main results of the paper being presented in Section~\ref{sec:results}. A summary of the results is captured in Figure~\ref{fig:summary} while the proofs are provided in Section~\ref{app:proofs}. Various definitions and notation that are used throughout the paper are collected in Section~\ref{sec:problem-formulation}.

\section{Definitions and Notation}\label{sec:problem-formulation}
$\N:=\{1,2,\ldots\}$ and $\R$ denotes the set of natural numbers and real numbers respectively. For $M\in \N$, $[M]:=\{1,\ldots,M\}$. $\b 1_d:=(1,1,\ldots,1)\in\R^d$ and $\b 0$ denotes the matrix of zeros. For $a:=(a_1,\ldots,a_d)\in\R^d$ and $b:=(b_1,\ldots,b_d)\in\R^d$, $\langle a,b\rangle=\sum^d_{i=1}a_ib_i$ is the Euclidean inner product. For sets $A$ and $B$, $A\backslash B=\{a\in A: a\notin B\}$ is their difference, $|A|$ is the cardinality of $A$ and $\times_{m=1}^M A_m =\{\left(a_1,\ldots,a_M\right):a_m\in A_m\, m\in[M]\}$ is the Descartes product of sets $(A_m)^M_{m=1}$. $\Po(\X)$ denotes the power set of a set $\X$, i.e., all subsets of $\X$ (including the empty set and $\X$). The Kronecker delta is defined as $\delta_{a,b} = 1$ if $a=b$, and zero otherwise. $\chi_A$ is the indicator function of set $A$: $\chi_A(x)=1$ if $x\in A$ and $\chi_A(x)=0$ otherwise. $\R^{d_1\times\ldots\times d_M}$ is the set of $d_1\times \ldots \times d_M$-sized tensors.

For a topological space $\left(\X,\tau_{\X}\right)$, $\B(\X):=\B(\tau_\X)$ is the Borel sigma-algebra on $\X$ induced by the topology $\tau_{\X}$. Probability and finite signed measures in the paper are meant w.r.t.\ the measurable space $(\X,\B(\X))$. Given $\left\{\left(\X_i,\tau_i\right)\right\}_{i\in I}$ topological spaces, their product $\times_{i\in I} \X_i$ is enriched with the product topology; it is the coarsest topology for which the canonical projections $\pi_i: \times_{i\in I} \X_i \rightarrow (\X_i,\tau_i)$ are 
continuous for all $i\in I$. A topological space $\left(\X,\tau_{\X}\right)$ is called second-countable if $\tau_{\X}$ has a countable basis.\footnote{Second-countability implies separability; in metric 
spaces the two notions coincide \cite[Proposition~2.1.4]{dudley04real}. By the Urysohn's theorem, a topological space is separable and metrizable if and only if it is regular, 
Hausdorff and second-countable. Any uncountable discrete space is \emph{not} second-countable.}
$C(\X)$ denotes the space of continuous functions on $\X$. $C_0(\X)$ denotes the class of real-valued functions vanishing at infinity on a locally compact Hausdorff (LCH) space\footnote{LCH spaces include $\R^d$, discrete spaces, and topological manifolds. Open or closed subsets,  finite products of LCH spaces are LCH. Infinite-dimensional Hilbert spaces are \emph{not} LCH.} $\X$, i.e., for any $\epsilon>0$, the set $\{x \in \X: |f(x)|\ge \epsilon\}$ is compact. $C_0(\X)$ is endowed with the uniform norm $\left\|f\right\|_{\infty} = \sup_{x\in \X} |f(x)|$. $\Mb(\X)$ and $\Mp(\X)$ are the space of finite signed measures and probability measures on $\X$, respectively.  For $\P_m\in \Mp(\X_m)$, $\otimes^M_{m=1}\P_m$ denotes the product probability measure on the product space $\times^M_{m=1}\X_m$, i.e., $\otimes^M_{m=1}\P_m\in \Mp(\times^M_{m=1}\X_m)$. 
$\delta_x$ is the Dirac measure supported on $x\in\X$. For $\F \in \Mb\left(\times_{m=1}^M \X_m\right)$, the finite signed measure $\F_m$ denotes its marginal on $\X_m$. $\H_{k_m}$ is the reproducing kernel Hilbert space (RKHS) associated with the reproducing kernel $k_m:\X_m\times\X_m\rightarrow\R$, which in this paper is assumed to be measurable and bounded. The tensor product of $(k_m)^M_{m=1}$ is a kernel, defined as
      \begin{align*}
      \otimes_{m=1}^M k_m\left(\left(x_1,\ldots,x_M\right),\left(x_1',\ldots,x_M'\right)\right) &= \prod_{m=1}^M k_m\left(x_m,x_m'\right), \quad x_m,x_m'\in \X_m,
      \end{align*}
whose associated RKHS is denoted as $\H_{\otimes_{m=1}^M k_m} = \otimes_{m=1}^M\H_{k_m}$ \citep[Theorem~13]{berlinet04reproducing}, where the r.h.s.\ is the tensor product of RKHSs $(\H_{k_m})^M_{m=1}$. For $h_m \in \H_m$, $m\in[M]$, the multi-linear operator $\otimes_{m=1}^M h_m \in \otimes_{m=1}^M \H_m$ is defined as
       \begin{align*}
          \left(\otimes_{m=1}^M h_m\right)\left(v_1,\ldots,v_M\right) = \prod_{m=1}^M \left<h_m,v_m\right>_{\H_m}, \quad v_m \in \H_m.
       \end{align*}
A kernel $k:\X \times \X \rightarrow \R$ defined on a LCH space $\X$ is called a $c_0$-kernel if $k(\cdot,x)\in C_0(\X)$ for all $x\in \X$. 
$k:\R^d \times \R^d \rightarrow \R$ is said to be a translation invariant kernel on $\R^d$ if $k(x, y) = \psi(x - y),\,x,y\in\R^d$ for a positive definite function $\psi: \R^d \rightarrow \R$. $\mu_k(\F)$ denotes the kernel mean embedding of $\F\in\Mb(\X)$ to $\H_k$ which is defined as $\mu_k(\F)=\int_{\X}k(\cdot,x)\,\d\F(x)$, where the integral is meant in the Bochner sense.

\section{Problem Formulation}\label{sec:problem}
In this section, we formally introduce the goal of the paper. To this end, we start with a definition. For simplicity, throughout the paper, we assume that all kernels are bounded. The definition is based on the observation \citep[Lemma 8]{sriperumbudur10hilbert} that a bounded kernel $k$ on a topological space $\left(\X,\tau_{\X}\right)$ is characteristic if and only if $$\int_{\X}\int_{\X}k(x,x')\,\d\F(x)\,\d\F(x')>0,\,\,\forall\,\F\in \Mb(\X)\backslash\{0\}\,\,\text{such that}\,\,\F(\X)=0.$$
In other words, characteristic kernels are integrally strictly positive definite (ispd; see \citealp[p. 1523]{sriperumbudur10hilbert}) w.r.t.~the class of finite signed measures that assign zero measure to $\X$. The following definition extends this observation to tensor product kernels on product spaces.
\begin{definition}[$\Fc$-ispd tensor product kernel]\label{def:F-char}
Suppose $k_m:\X_m \times \X_m \rightarrow \R$ is a bounded kernel on a topological space $\left(\X_m,\tau_{\X_m}\right),\,m\in[M]$. Let $\Fc \subseteq \Mb\left( \X \right)$ be such that $0\in \Fc$ where $\X:=\times^M_{m=1}\X_m$.
 $k:=\otimes_{m=1}^M k_m$ is said to be \emph{$\Fc$-ispd} if
 \begin{gather}
  \mu_k(\F) = 0 \Rightarrow \F = 0 \quad (\F \in \Fc)\text{, or equivalently}\nonumber\\
   \left\|\mu_k(\F)\right\|_{\H_k}^2 = \int_{\times_{m=1}^M\X_m}\int_{\times_{m=1}^M\X_m} \left(\otimes_{m=1}^M k_m\right)\left(x,x'\right)\,\d \F(x)\, \d \F(x')>0, \quad \forall\, \F \in \Fc \backslash \{0\}. \label{def:F-char>0}
 \end{gather}
 Specifically,\vspace{1mm}
 \begin{compactitem}
   \item if $k_m$-s are $c_0$-kernels on locally compact Polish (LCP)~\footnote{A topological space is called Polish if it is complete, separable and metrizable. For example, $\R^d$ and countable discrete spaces are Polish. Open and closed subsets, products and disjoint unions of countably many Polish spaces are Polish. Every second-countable LCH space is Polish.} spaces $\X_m$-s and $\Fc = \Mb(\X)$, then $k$ is called $c_0$-universal.\vspace{1mm}
   \item if
	 \begin{align*}
	   \Fc &= \left[\Mb(\X)\right]^0\hspace{1.35cm} := \left\{\F \in \Mb(\X): \F(\X)=0\right\},\\
\Fc &= \left[\otimes_{m=1}^M\Mb(\X_m)\right]^0 := \left\{ \F \in \otimes_{m=1}^M\Mb\left(\X_m\right), \F(\X) = 0 \right\},\\
	   \Fc &= \I \hspace{2.6cm} := \left\{\P - \otimes_{m=1}^M \P_m: \P \in \Mp\left(\times_{m=1}^M\X_m\right)\right\}, \quad (M\ge 2)\\
	   \Fc &= \otimes_{m=1}^M \Mb^0(\X_m)\hspace{0.48cm}:=\left\{ \F = \otimes_{m=1}^M \F_m\,:\, \F_m\in\Mb\left(\X_m\right),\,\F_m(\X_m) = 0,\,\, \forall\, m\in[M]\right\},
	 \end{align*}
	  then $k$ is called \emph{characteristic}, \emph{$\otimes$-characteristic}, \emph{$\I$-characteristic} and \emph{$\otimes_0$-characteristic}, respectively.
 \end{compactitem}
\end{definition}
In Definition~\ref{def:F-char}, $k$ being characteristic matches the usual notion of characteristic kernels on a product space, i.e., there are no two distinct probability measures 
on $\X=\times^M_{m=1}\X_m$ such that the MMD between them is zero. The other notions such as $\otimes$-characteristic, $\I$-characteristic and $\otimes_0$-characteristic are typically weaker than the usual characteristic property since
\begin{equation}\label{Eq:measure-subset}
			\xymatrixcolsep{0.5cm} \xymatrixrowsep{0.4cm}
			 \xymatrix{
			   \otimes_{m=1}^M \Mb^0(\X_m) \ar@{}[r]|{\subseteq} &\left[\otimes_{m=1}^M\Mb(\X_m)\right]^0 \ar@{}[r]|{\subseteq} & \left[\Mb\left(\times_{m=1}^M\X_m\right)\right]^0 \ar@{}[r]|{\subseteq} & \Mb\left(\times_{m=1}^M \X_m\right)\\
			   & & \I \ar@{}[u]|{\vsubseteq} &
			 }.
\end{equation}
Below we provide further intuition on the $\Fc$ measure classes enlisted in Definition~\ref{def:F-char}.%\vspace{2mm}

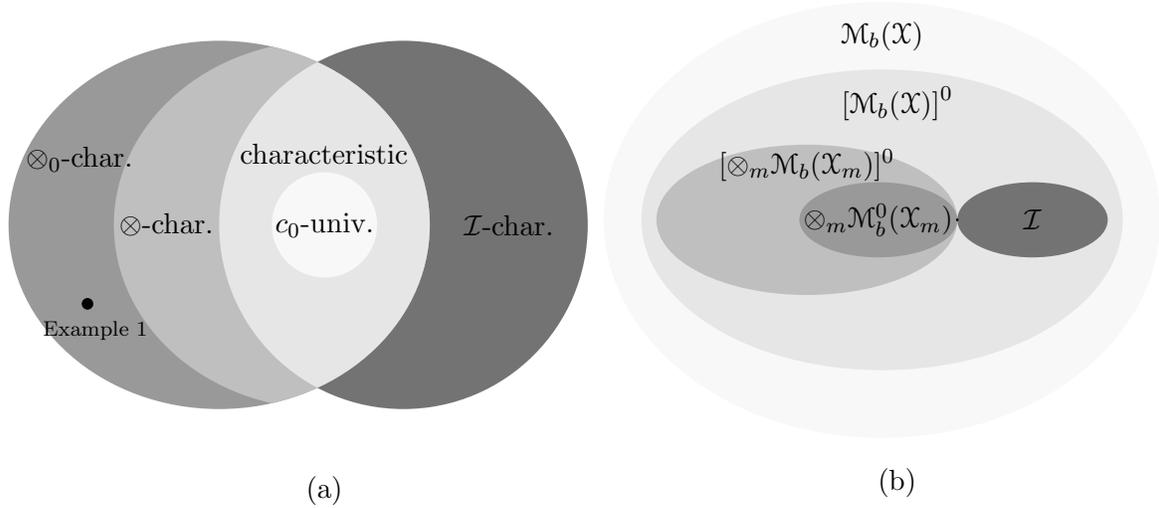
\begin{figure}[t]
\raisebox{-0.1cm}{
\begin{tikzpicture}[scale = 0.35]
   \def\firstellipse{(-3,0) ellipse (8 and 7)}
   \def\secondellipse{(4,0) ellipse (7 and 7)}
   \def\thirdellipse{(1,0) ellipse (8 and 7)}

  \begin{scope}
  % colour ellipses
  \fill[gray4] \firstellipse;
  \fill[gray5] \secondellipse;
  \end{scope}

  % colour intersection
   \begin{scope}
     \clip \firstellipse;
     \fill[gray3] \thirdellipse;
   \end{scope}

   % colour intersection
   \begin{scope}
     \clip \firstellipse;
     \fill[gray2] \secondellipse;
   \end{scope}

   \fill[fill=gray1] (1,0) circle (2);

   \node at (1,0) (A) {$c_0$-univ.};
   \node at (1,2.7) (B) {characteristic};
   \node at (8,0) (C) {$\I$-char.};
   \node at (-5,0) (D) {$\otimes$-char.};
   \node at (-8.3,2.5) (E) {$\otimes_0$-char.};
   \node at (1,-10.1) (F) {(a)};
   
     %Example-1:
  \draw[fill=black] (-8,-3) circle [radius = 0.2];	
  \node at (-7.7,-4) (G) {\scriptsize Example~\ref{example1:k_m:char=does not=>k_1xk_2:char-or-x-char,but-here-I-char}};
\end{tikzpicture}
}
\begin{tikzpicture}
  \begin{scope}[scale=1]%,fill opacity=0.5
	%\draw[fill=gray1] (-1,0) circle (4);
	%c_0-universal:
	\fill[fill=gray1] (-1,0) ellipse (3.7 and 2.9);
	\node at (-1,2.45) (E) {$\Mb(\X)$};
	
	%char
	\fill[fill=gray2] (-1,0) ellipse (3.2 and 2);
	\node at (-0.8,1.55) (D) {$\left[\Mb(\X)\right]^0$};
	
	%x-char:
	\fill[fill=gray3] (-2,0) ellipse (2 and 1);
	
	%x_0-char:
	\fill[fill=gray4] (-1.05,0) ellipse (1.05 and 0.5);
	
	%I-char
	\fill[fill=gray5] (1,0) ellipse (1 and 0.5);
	
	\node at (1,0) (A) {$\I$};
	\node at (-1.05,0) (B) {$\otimes_m \Mb^0(\X_m)$};
	\node at (-2,0.73) (C) {$[\otimes_m\Mb(\X_m)]^0$};
	
	\draw[fill=black] (0,0) circle [radius = 0.01];
	\node at (-0.8,-3.5) (F) {(b)};
  \end{scope}

\end{tikzpicture}
\caption{(a) $\Fc$-ispd $\otimes_{m=1}^Mk_m$ kernels (see \eqref{eq:char-relations}); (b) $\Fc \subseteq \Mb(\X)$, $\X=\times_{m=1}^M\X_m$. 
Example~\ref{example1:k_m:char=does not=>k_1xk_2:char-or-x-char,but-here-I-char}: $\otimes_{m=1}^Mk_m$ is $\otimes_0$-characteristic but not $\otimes$-characteristic and therefore not 
characteristic.\label{fig:F-char-demo}}
\vspace{-4mm}
\end{figure}

\begin{remark}
\begin{itemize}[labelindent=0.6em,leftmargin=*,topsep=0cm,partopsep=0cm,parsep=0cm,itemsep=0cm]
    \item[(i)] $\bm{\Fc=\Mb(\X):}$ If  $k_m$-s are $c_0$-kernels on LCH spaces $\X_m$ for all $m\in[M]$, then $k$ is also a $c_0$-kernel on LCH space $\X$ implying that if $k$ satisfies \eqref{def:F-char>0}, then $k$ is $c_0$-universal (\citealp[Proposition~2]{sriperumbudur10hilbert}). It is well known \citep{sriperumbudur10hilbert} that $c_0$-universality reduces to 
    $c$-universality (i.e., the notion of universality proposed by \citealp{steinwart01influence}) if $\X$ is compact which is guaranteed if and only if each $\X_m,\,m\in[M]$ is compact.\vspace{2mm}
    \item[(ii)] $\bm{\Fc = \I:}$ This  family  is useful to describe the joint \emph{independence} of $M$ random variables---hence the name $\I$-characteristic---defined on kernel-endowed 
    domains $(\X_m)_{m=1}^M$: If $\P$ denotes the joint distribution of random variables $(X_m)_{m=1}^M$ and $(\P_m)_{m=1}^M$ are the associated marginals on $(\X_m)_{m=1}^M$, then by definition $k=\otimes_{m=1}^M k_m$ is $\I$-characteristic iff
	      \begin{align*}
		\emph{HSIC}_k(\P)=0\Longleftrightarrow \P=\otimes^M_{m=1}\P_m.
	      \end{align*}
	    In other words, HSIC captures joint independence exactly with $\I$-characteristic kernels.
	    Similarly, the $\I$-characteristic property ensures that COCO (constrained covariance; \citealp{gretton05kernel}) is a joint independence measure as COCO is defined by replacing the Hilbert-Schmidt norm of the cross-covariance operator (see \eqref{eq:def:HSIC:M=2} and \eqref{eq:def:HSIC:M>=2}) with its spectral norm.\vspace{2mm}
    \item[(iii)] $\bm{\Fc = \otimes_{m=1}^M \Mb^0(\X_m):}$ In this case  $\Fc$ is chosen to be the product of finite signed measures on $\X$ such that 
    each marginal measure $\F_m$ assigns zero to the corresponding space $\X_m$. This choice is relevant as the characteristic property of individual kernels $(k_m)^M_{m=1}$ need \emph{not} imply the characteristic property of $\otimes_{m=1}^M k_m$, but is equivalent to the 
    $\otimes_0$-characteristic property of $\otimes_{m=1}^M k_m$.  The equivalence holds for bounded kernels $k_m: \X_m \times \X_m \rightarrow \R$ on topological spaces $\X_m$ ($m\in[M]$) since
      for any $\F = \otimes_{m=1}^M \F_m \in \otimes_{m=1}^M\Mb\left(\X_m\right)$, $\F_m(\X_m) = 0$ ($\forall\, m\in[M])$ 
      \begin{align}
		\left\|\mu_k(\F)\right\|_{\H_{\otimes_{m=1}^M k_m}}^2 &= \prod_{m=1}^M\left\|\mu_{k_m}(\F_m)\right\|^2_{\H_{k_m}},\label{Eq:prod}
	      \end{align} and the l.h.s.\ is positive iff each term on the r.h.s.~is positive.\vspace{2mm}   
\item[(iv)] $\bm{\Fc = \left[\otimes_{m=1}^M \Mb(\X_m)\right]^0:}$ This class is similar to the one discussed in \textit{(iii)} above---i.e., class of product measures---with the slight difference that the joint measure $\F$ is restricted to assign zero measure to $\X$ without requiring all the marginal measures $\F_m$ to assign zero measure 
to the corresponding space $\X_m$. While the need for considering such
a measure class may not be clear at this juncture, however, based on \eqref{Eq:prod}, it turns out that this choice of $\Fc$ has quite surprising connections to the characteristic property 
and $c_0$-universality of the product kernel; for details see Remark~\ref{Rem:new}.
\vspace{2mm}
    \item[(v)] \tb{$\bm{\Fc}$-ispd relations:} Given the relations in \eqref{Eq:measure-subset}, it immediately follows that $k=\otimes^M_{m=1}k_m$ satisfies
	  \begin{equation}
		    \xymatrixcolsep{0.63cm} \xymatrixrowsep{0.4cm}
		    \hspace*{-1cm}
		    \xymatrix{
				  \otimes_0\text{-characteristic} & \ar@{}[l]|{\Longleftarrow}  \otimes\text{-characteristic} &\ar@{}[l]|{\Longleftarrow} \ar@{}[d]|{\rotatebox{90}{\text{$\Leftarrow$}}} \text{characteristic} & \ar@{}[l]|{\Longleftarrow} c_0\text{-universal}\\
				   & & \I\text{-characteristic}
		    }\label{eq:char-relations}
		    \end{equation}
	    when $\X_m$ for all $m\in[M]$ are LCP. A visual illustration of \eqref{Eq:measure-subset} and \eqref{eq:char-relations} is provided in Figure~\ref{fig:F-char-demo}. \vspace{2mm}
\item[(vi)] $\bm{\left[\otimes_{m=1}^M \Mb(\X_m)\right]^0 \cap \I =\{0\}:}$ While it is clear that $\left[\otimes_{m=1}^M \Mb(\X_m)\right]^0$ and $\I$ are subsets of $\left[\Mb(\times^M_{m=1}\X_m)\right]^0$, 
it is interesting to note that $\left[\otimes_{m=1}^M \Mb(\X_m)\right]^0$ and $\I$ have a trivial intersection with $0$ being the measure common to each of them, assuming that 
$\X_m$-s are second-countable for all $m\in[M]$; see Section~\ref{subsec:0-intersection}.
\end{itemize}
\label{Rem:remark}
\end{remark}
Having defined the $\Fc$-ispd property, our \tb{goal} is to 
		    investigate whether the characteristic or $c_0$-universal property of $k_m$-s ($m\in[M]$) imply different $\Fc$-ispd properties of $\otimes_{m=1}^M k_m$, and vice versa.

\section{Main Results}\label{sec:results}
In this section, we present our main results related to the $\Fc$-ispd property of tensor product kernels, which are summarized in Figure~\ref{fig:summary}. 
The results in this section will deal with various assumptions on $\X_m$, such as second-countability, Hausdorff, locally compact Hausdorff (LCH) and locally compact Polish (LCP), so that they are presented in more generality. However, for simplicity, all these assumptions can be unified by simply assuming a stronger condition that $\X_m$'s are LCP.

Our first example illustrates that the characteristic property of $k_m$-s does not imply the characteristic property of the tensor product kernel. In light of Remark~\ref{Rem:remark}(iv) of Section~\ref{sec:problem}, it follows that the class of $\otimes_0$-characteristic tensor product kernels form a 
\emph{strictly} larger class than  characteristic tensor product kernels; see also Figure~\ref{fig:F-char-demo}.
\begin{example}
\label{example1:k_m:char=does not=>k_1xk_2:char-or-x-char,but-here-I-char}
  Let $\X_1 = \X_2 = \{1,2\}$, $\tau_{\X_1} = \tau_{\X_2} = \Po(\{1,2\})$, $k_1(x,x') = k_2(x,x') = 2 \delta_{x,x'} - 1$. It is easy to verify that
  $k_1$ and $k_2$ are characteristic. However, it can be proved that $k_1\otimes k_2$ is \emph{not} $\otimes$-characteristic and therefore not characteristic. On the hand, interestingly, $k_1\otimes k_2$ is $\I$-characteristic. We refer the reader to Section~\ref{subsec:ex1} for details.
\end{example}

In the above example, we showed that the tensor product of $k_1$ and $k_2$ (which are characteristic kernels) is $\I$-characteristic.
The following result generalizes this behavior for any bounded characteristic kernels. In addition, under a mild assumption, it shows the converse to be true for any $M$.

\begin{theorem}
\label{thm2:k_m:char<=>xk_m:I-char}
Let $k_m: \X_m \times \X_m \rightarrow \R$ be bounded kernels on topological spaces $\X_m$ for all $m\in[M]$, $M\ge 2$. Then the following holds.\vspace{1mm}
 \begin{compactitem}
 \item[(i)] Suppose $\X_m$ is second-countable for all $m\in[M]$ with $M=2$. If $k_1$ and $k_2$ are characteristic, then $k_1 \otimes k_2$ is $\I$-characteristic.\vspace{.5mm}
 \item[(ii)] Suppose $\X_m$ is Hausdorff and $|\X_m|\ge 2$ for all $m\in[M]$. If $\otimes_{m=1}^M k_m$ is $\I$-characteristic, then $k_1,\ldots,k_M$ are characteristic.
\end{compactitem}
\end{theorem}

\citet{lyons13distance} has showed an analogous result to Theorem~\ref{thm2:k_m:char<=>xk_m:I-char}(i) for distance covariances ($M=2$) on metric spaces of negative type (\citealp[Theorem~3.11]{lyons13distance}), which
by \citet[Proposition~29]{sejdinovic13equivalence} holds for HSIC yielding the $\I$-characteristic property of $k_1\otimes k_2$.
Recently, \citet{gretton15simpler} presented a direct proof showing that HSIC corresponding to $k_1 \otimes k_2$ captures independence if $k_1$ and $k_2$ are translation invariant characteristic kernels on $\R^d$ (which is equivalent to $c_0$-universality). 
\citet{blanchard11generalizing} proved a result similar to Theorem~\ref{thm2:k_m:char<=>xk_m:I-char}(i) assuming that $\X_m$'s are compact and $k_1$, $k_2$ being $c$-universal. In contrast, Theorem~\ref{thm2:k_m:char<=>xk_m:I-char}(i) establishes the result for bounded kernels on 
general second-countable topological spaces. In fact, the results of \citet{gretton15simpler,blanchard11generalizing} are special cases of Theorems~\ref{thm4:contboundedshiftinv} and \ref{thm4:prod-of-c0universal} below. Theorem~\ref{thm2:k_m:char<=>xk_m:I-char}(i) raises a pertinent question: whether $\otimes^M_{m=1}k_m$ is $\I$-characteristic if $k_m$-s are characteristic for all $m\in[M]$ where $M>2$? The following example provides a negative answer to this question.
On a positive side, however, we will see in Theorem~\ref{thm4:prod-of-c0universal} that the $\I$-characteristic property of $\otimes_{m=1}^M k_m$ can be guaranteed for any
$M\ge 2$ if a stronger condition is imposed on $k_m$-s (and $\X_m$-s). Theorem~\ref{thm2:k_m:char<=>xk_m:I-char}(ii) generalizes Proposition~3.15 of \citet{lyons13distance} for any $M>2$, which states that every kernel $k_m,\,m\in[M]$ being characteristic is necessary for the tensor kernel $\otimes^M_{m=1}k_m$ to be $\I$-characteristic.

\begin{example}
\label{example:k_m:char=does not=>x_m k_m:I-char}
Let $M=3$ and $\X_m := \{1,2\}$, $\tau_{\X_m}=\Po(\X_m)$, $k_m\left(x,x'\right) = 2 \delta_{x,x'} - 1$ ($m=1,2,3$). 
As mentioned in Example~\ref{example1:k_m:char=does not=>k_1xk_2:char-or-x-char,but-here-I-char}, $(k_m)^3_{m=1}$ are characteristic. However, it can be shown that $\otimes_{m=1}^3{k_m}$ is not $\I$-characteristic. See Section~\ref{subsec:ex2} for details.
\end{example}

In Remark~\ref{Rem:remark}(iii) and Example~\ref{example1:k_m:char=does not=>k_1xk_2:char-or-x-char,but-here-I-char},
we showed that in general, only the $\otimes_0$-characteristic property of $\otimes_{m=1}^M k_m$ is equivalent to the characteristic property of $k_m$-s.
Our next result shows that all the various notions of characteristic property of $\otimes_{m=1}^M k_m$ \emph{coincide} if $k_m$-s are translation-invariant, continuous bounded kernels on $\R^d$.

\begin{theorem}
\label{thm4:contboundedshiftinv}
Suppose $k_m: \R^{d_m} \times \R^{d_m} \rightarrow \R$ are continuous, bounded and translation-invariant kernels for all $m\in[M]$.
Then the following statements are equivalent:\vspace{1mm}
  \begin{compactenum}
	\item[(i)] $k_m$-s are characteristic for all $m\in[M]$;\vspace{.5mm}
	\item[(ii)] $\otimes_{m=1}^M k_m$ is $\otimes_0$-characteristic;\vspace{.5mm}
	\item[(iii)] $\otimes_{m=1}^M k_m$ is $\otimes$-characteristic;\vspace{.5mm}
	\item[(iv)] $\otimes_{m=1}^M k_m$ is $\I$-characteristic;\vspace{.5mm}
	\item[(v)] $\otimes_{m=1}^M k_m$ is characteristic.
  \end{compactenum}
\end{theorem}

The following result shows that on LCP spaces, the tensor product of $M\ge 2$ $c_0$-universal kernels is also $c_0$-universal, and vice versa.

\begin{theorem}
\label{thm4:prod-of-c0universal}
Suppose $k_m: \X_m \times \X_m \rightarrow \R$ are $c_0$-kernels on LCP spaces $\X_m$ ($m\in [M]$).
  Then $\otimes_{m=1}^Mk_m$ is $c_0$-universal iff $k_m$-s are $c_0$-universal for all $m\in[M]$.
\end{theorem}
\begin{remark}\label{rem:Remark2}
\begin{itemize}[labelindent=0.6em,leftmargin=*,topsep=0cm,partopsep=0cm,parsep=0cm,itemsep=2mm]
    \item[(i)] 
A special case of Theorem~\ref{thm4:prod-of-c0universal} for $M=2$ is proved by
	  \citet[Lemma 3.8]{lyons13distance} in the context of distance covariance which reduces to Theorem~\ref{thm4:prod-of-c0universal} through the equivalence established by \citet{sejdinovic13equivalence}. 
	  Another special case of Theorem~\ref{thm4:prod-of-c0universal} is proved by \citet[Lemma~5.2]{blanchard11generalizing} for $c$-universality with $M=2$ using the Stone-Weierstrass theorem: if $k_1$ and $k_2$ are $c$-universal 
	 then $k_1\otimes k_2$ is $c$-universal.
    \item[(ii)] 
Since the notions of $c_0$-universality and characteristic property are equivalent for translation invariant $c_0$-kernels on $\R^d$ (\citealp[Prop.~5.16]{carmeli10vector}, 
	 \citealp[Theorem~9]{sriperumbudur10hilbert}), Theorem~\ref{thm4:contboundedshiftinv} can be considered as a special case of 
	 Theorem~\ref{thm4:prod-of-c0universal}. In other words, requiring $(k_m)_{m=1}^M$ to be also $c_0$-kernels in Theorem~\ref{thm4:contboundedshiftinv}(i)-(iv) is equivalent to
	 \vspace{1mm}
	    \begin{compactenum}
		\item[(v)] $k_m$-s are $c_0$-universal for all $m\in[M]$;
		\item[(vi)] $\otimes_{m=1}^M k_m$ is $c_0$-universal.
	  \end{compactenum}
    \item[(iii)] 
Since the $c_0$-universality of $\otimes_{m=1}^Mk_m$ implies its $\I$-characteristic property (see \eqref{eq:char-relations}), Theorem~\ref{thm4:prod-of-c0universal} also provides a 
	 generalization of Theorem~\ref{thm2:k_m:char<=>xk_m:I-char}(i) to $M\ge 2$ under additional assumptions on $k_m$-s, while constraining $\X_m$-s to LCP-s instead of second-countable topological spaces.
\end{itemize}
\end{remark}

In Example~\ref{example:k_m:char=does not=>x_m k_m:I-char} and Theorem~\ref{thm4:prod-of-c0universal}, we showed that for $M\ge 3$ components while the characteristic property of $(k_m)_{m=1}^M$ is not sufficient, their universality is enough to guarantee the $\I$-characteristic property of $\otimes_{m=1}^Mk_m$. The next example demonstrates that these results are tight:
If at least one $k_m$ is not universal but only characteristic, then $\otimes_{m=1}^Mk_m$ might not be $\I$-characteristic.
\begin{example} \label{example:k1,k2:univ,k_3:char=does not=>x_m k_m:I-char}
Let $M=3$ and $\X_m := \{1,2\}$, $\tau_{\X_m}=\Po(\X_m)$, for all $m\in [3]$, $k_1\left(x,x'\right) = 2 \delta_{x,x'} - 1$, and $k_m\left(x,x'\right) = \delta_{x,x'}$  ($m=2,3$). 
$k_1$ is characteristic (Example~\ref{example1:k_m:char=does not=>k_1xk_2:char-or-x-char,but-here-I-char}), $k_2$ and $k_3$ are universal since the associated Gram matrix 
$\b{G} = [k_m(x,x')]_{x,x'\in \X_m}$ is an identity matrix, which is strictly positive definite ($m=2,3$).
However, $\otimes_{m=1}^3{k_m}$ is \emph{not} $\I$-characteristic. See Section~\ref{subsec:ex3} for details.
\end{example}

\begin{figure}[t]
\begin{center}
\begin{tikzpicture}[scale = 0.35]
   \def\firstellipse{(-2.5,0) ellipse (7 and 7)}
   \def\secondellipse{(3.5,0) ellipse (7 and 7)}
   \def\thirdellipse{(1,0) ellipse (8 and 7)}

   \begin{scope}
   % colour ellipses
   \fill[gray4] \firstellipse;
   \fill[gray5] \secondellipse;
   \end{scope}

    % colour intersection
    \begin{scope}
      \clip \firstellipse;
      \fill[gray2] \secondellipse;
    \end{scope}

   %=$c_0$-universal = $\otimes$-char.
   \node at (0.5,2.8) (B) {$\otimes$-char.};
   \node at (0.5,1) (B2) {=characteristic};
   \node at (0.5,-0.8) (B3) {=$c_0$-universal};
   \node at (8,0) (C) {$\I$-char.};
   \node at (-6.5,0) (E) {$\otimes_0$-char.};
\end{tikzpicture}
\end{center}
\caption{Simplification of the $\Fc$-ispd property of tensor product kernels; see Remark~\ref{Rem:new}.\label{fig:F-char-demo1}}
\end{figure}
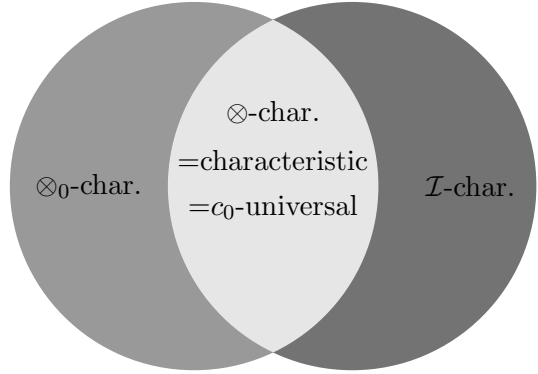

\begin{remark}\label{Rem:new}
Note that the l.h.s.~in \eqref{Eq:prod} is positive if and only if each term on the r.h.s.~is positive, i.e., if $k=\otimes^M_{m=1}k_m$ is $\otimes$-characteristic with 
$k_m$-s being $c_0$-kernels on LCP $\X_m$-s, then all $k_m$-s are $c_0$-universal. A similar result was also proved by \citet[Lemma 3.4]{steinwart-ziegel}. Combining this with Theorem~\ref{thm4:prod-of-c0universal} yields that for tensor product $c_0$-kernels, the notions of $\otimes$-characteristic, characteristic and $c_0$-universality are equivalent, which is quite surprising as for a joint kernel $k$ (that is not of product type), these notions need not necessarily coincide. In light of this discussion, Figure~\ref{fig:F-char-demo}(a) can be simplified to Figure~\ref{fig:F-char-demo1}.
\end{remark}

\section{Proofs} \label{app:proofs}
In this section, we provide the proofs of our results presented in Section~\ref{sec:results}.
\subsection{Proof of Remark~\ref{Rem:remark}(iv)} \label{subsec:0-intersection}
By the second-countability of $\X_m$-s, $\B\left(\times_{m=1}^M \X_m\right)=\otimes_{m=1}^M \B(\X_m)$, where the r.h.s. is defined as the $\sigma$-field generated by the cylinder sets 
$A_m \times_{n\ne m} \X_n$ where $m\in [M]$ and $A_m \in \B(\X_m)$. Suppose there exists $\F \in \left[\otimes_{m=1}^M \Mb(\X_m)\right]^0 \cap \I$ such that $\F\ne 0$. This means there exists $\P\in \Mp\left(\times_{m=1}^M\X_m\right)$ with $(\P_m)_{m=1}^M$ being the marginals of $\P$ such that $\F=\otimes_{m=1}^M \F_m = \P-\otimes_{m=1}^M\P_m$. Since $\F \ne 0$ there exists  $A_m \times_{n\ne m} \X_n$ for some $m\in [M]$ and  $A_m \in \B(\X_m)$ such that $0\ne \F(A_m \times_{n\ne m} \X_n) = \F_m(A_m) \prod_{n\ne m} \F_n(\X_n) = \P\left(A_m \times_{n\ne m}\X_n\right) - \P_m(A_m) \prod_{n\ne m}\P_n(\X_n) = \P_m(A_m) - \P_m(A_m)=0$, leading to a contradiction.

\subsection{Proof of Example~\ref{example1:k_m:char=does not=>k_1xk_2:char-or-x-char,but-here-I-char}}\label{subsec:ex1}
The proof is structured as follows.\vspace{1mm}
  \begin{compactenum}
	\item First we show that $k:=k_1 = k_2$ is a kernel and it is characteristic.\vspace{.5mm}
	\item Next it is proved that $k_1 \otimes k_2$ is not $\otimes$-characteristic, which implies $k_1\otimes k_2$ is not characteristic.\vspace{.5mm}
	\item Finally, the $\I$-characteristic property of $k_1\otimes k_2$ is established. \vspace{2mm}
  \end{compactenum}
  The individual steps are as follows:\vspace{2mm}\\
\noindent \tb{$k$ is a kernel.} Assume w.l.o.g.\ that $x_1 = \ldots = x_N = 1$, $x_{N+1} = \ldots = x_{n} = 2$. Then it is easy to verify that the Gram matrix $\b G = [k(x_i,x_j)]_{i,j=1}^n=\b a \b a^\top$ where $\b a:=\left(\b 1^\top_{N},-\b 1^\top_{n-N}\right)^\top$ and $\b a^\top$ is the transpose of $\b a$. Clearly $\b G$ is positive semidefinite and so $k$ is a kernel. \vspace{2mm}\\
\noindent\tb{$k$ is characteristic}. We will show that $k$ satisfies \eqref{def:F-char>0}.
On $\X=\{1,2\}$ a finite signed measure $\F$ takes the form $\F = a_1\delta_1 + a_2\delta_2$ for some $a_1, a_2\in \R$. Thus,
\begin{equation}\label{Eq:tt}
\F \in \Mb(\X)\backslash\{0\}  \Leftrightarrow (a_1,a_2)\ne \b 0\quad\quad\text{and}\quad\quad\F(\X) = 0 \Leftrightarrow a_1 + a_2 = 0.
\end{equation}
Consider
		\begin{align}
		 \int_{\X}\int_{\X} k(x,x')\,\d \F(x)\, \d \F(x') &= a_1^2 k(1,1) + a_2^2 k(2,2) + 2 a_1 a_2 k(1,2)\nonumber\\
        & = a_1^2 + a_2^2 -2 a_1 a_2 = (a_1 -a_2)^2 = 4 a_1^2 >0,\label{Eq:char-univ}
		\end{align}
		where we used \eqref{Eq:tt} and the facts that $k(1,1)=k(2,2)=1$, $k(1,2)=-1$. \vspace{2mm}\\
\noindent \tb{$k_1 \otimes k_2$ is not $\otimes$-characteristic}. 
		   We  construct a witness $\F = \F_1 \otimes \F_2 \in \otimes_{m=1}^2 \Mb(\X_m) \backslash \{0\}$ such that
		   \begin{align}
			\F(\X_1 \times \X_2)&=\F_1(\X_1) \F_2(\X_2)=0, \label{eq:not-x-char:1}
\end{align}
and
\begin{align}
			 0 & = \int_{\X_1 \times \X_2} \int_{\X_1 \times \X_2} \underbrace{(k_1 \otimes k_2)((i_1,i_2),(i_1',i_2'))}_{k_1(i_1,i_1')k_2(i_2,i_2')}\, \d \F(i_1,i_2)\, \d \F(i_1',i_2')\nonumber \\
			 & = \prod_{m=1}^2 \int_{\X_m}\int_{\X_m} k_m(i_m,i_m')\, \d\F_m(i_m)\, \d \F_m(i_m'). \label{eq:not-x-char:2}
			\end{align}
			Finite signed measures on $\{1,2\}$ take the form $\F_1 =\F_1(\b a) = a_1 \delta_1 + a_2 \delta_2$, $\F_2 = \F_2(\b b) = b_1 \delta_1 + b_2 \delta_2$ form, where $\b a =(a_1,a_2)\in\R^2, \b b = (b_1,b_2) \in\R^2$.
			With these notations, \eqref{eq:not-x-char:1} and \eqref{eq:not-x-char:2} can be rewritten as
			\begin{align*}
						0&= (a_1+a_2)(b_1+b_2),\\
						0 &= \left[\sum_{i,i'=1}^2 k_1(i,i')a_{i}a_{i'}\right] \left[ \sum_{j,j'=1}^2 k_2(j,j')b_{j}b_{j'} \right] = (a_1-a_2)^2(b_1-b_2)^2.
			\end{align*}
			Keeping the solutions where neither $\b a$ nor $\b b$ is the zero vector, there are 2 (symmetric) possibilities: (i) $a_1 + a_2 =0$, $b_1=b_2$ and (ii) $a_1=a_2$, $b_1+b_2=0$. In other words, for any $a,b\ne 0$, the possibilities are (i) $\b a=(a,-a)$, $\b b =(b,b)$ and (ii) $\b a=(a,a)$, $\b b=(b,-b)$.
			This establishes the non-$\left[\otimes_{m=1}^2 \Mb(\X_m)\right]^0$-ispd property of $k_1\otimes k_2$. \vspace{2mm}\\
\noindent \tb{$k_1 \otimes k_2$ is $\I$-characteristic}.
		   Our goal is to show that $k_1\otimes k_2$ is $\I$-characteristic, i.e., for any $\P\in \Mp(\X_1 \times \X_2)$, $\mu_{k_1\otimes k_2}(\F) = 0$ implies $\F = 0$,
		   where $\F = \P - \P_1 \otimes \P_2$. We divide the proof into two parts: \vspace{1mm}
		   \begin{compactenum}
			 \item First we derive the equations of
			  \begin{align}
				\F(\X_1 \times \X_2) &= 0 \quad \text{and} \quad
			   \int\int_{(\X_1 \times \X_2)^2} (k_1\otimes k_2)\left((i,j),(r,s)\right)\, \d \F(i,j)\, \d \F(r,s) = 0
\label{eq:I-char:2a}
			  \end{align}			
			  for general finite signed measures $\F = \sum_{i,j=1}^2 a_{ij} \delta_{(i,j)}$ on $\X_1 \times \X_2$.\vspace{.5mm}
			 \item Then, we apply the $\F = \P - \P_1 \otimes \P_2$ parameterization and solve for $\P$ that satisfies \eqref{eq:I-char:2a} to conclude that $\P=\P_1\otimes \P_2$, i.e., $\F=0$. Note that in the chosen parametrization for $\F$, $\F(\X_1 \times \X_2) = 0$ holds  automatically.\vspace{2mm}
		   \end{compactenum}
\noindent		   The details are as follows. \vspace{2mm}\\
\noindent \tb{Step 1}.
				   \begin{align}
						0 &= \F(\X_1 \times \X_2) \Leftrightarrow 0 = a_{11} + a_{12} + a_{21} + a_{22}, \label{eq:FXY}\\
						0 &= \int_{\X_1 \times \X_2}\int_{\X_1 \times \X_2} \underbrace{(k_1\otimes k_2)\left((i,j),(r,s)\right)}_{k_1(i,r)k_2(j,s)}\,\d \F(i,j)\, \d \F\left(r,s\right) \nonumber\\
						&= \sum_{i,j=1}^2 \sum_{r,s=1}^2 k_1(i,r)k_2(j,s)a_{ij}a_{rs} = \sum_{i,r=1}^2 k_1(i,r) \sum_{j,s=1}^2 k_2(j,s)a_{ij}a_{rs}\nonumber \\
						&= k_1(1,1)\left[k_2(1,1) a_{11} a_{11} + k_2(1,2) a_{11} a_{12} + k_2(2,1) a_{12} a_{11} + k_2(2,2) a_{12} a_{12}\right]  \nonumber\\
						& \quad +k_1(1,2)\left[k_2(1,1) a_{11} a_{21} + k_2(1,2) a_{11} a_{22} + k_2(2,1) a_{12} a_{21} + k_2(2,2) a_{12} a_{22} \right] \nonumber\\
						& \quad +k_1(2,1)\left[k_2(1,1) a_{21} a_{11} + k_2(1,2) a_{21} a_{12} + k_2(2,1) a_{22} a_{11} + k_2(2,2) a_{22} a_{12} \right] \nonumber\\
						& \quad +k_1(2,2)\left[k_2(1,1) a_{21} a_{21} + k_2(1,2) a_{21} a_{22} + k_2(2,1) a_{22} a_{21} + k_2(2,2) a_{22} a_{22} \right]\nonumber\\
						&= \underbrace{\left(a_{11}^2 - 2a_{11}a_{12} + a_{12}^2\right)}_{\left(a_{11} -a_{12}\right)^2} +
						 \underbrace{\left(a_{21}^2-2a_{21}a_{22} + a_{22}^2\right)}_{\left(a_{21} - a_{22}\right)^2}
- 2 \underbrace{\left(a_{11}a_{21}-a_{11}a_{22}-a_{12}a_{21}+a_{12}a_{22}\right)}_{(a_{11}-a_{12}) (a_{21}-a_{22})} \nonumber\\
					   & = (a_{11} - a_{12} - a_{21} + a_{22})^2. \label{eq:FQuadr}
					\end{align}
					Solving \eqref{eq:FXY} and \eqref{eq:FQuadr} yields
				   \begin{align}
					  a_{11} + a_{22} &= 0\quad\text{and}\quad a_{12} + a_{21} = 0. \label{eq:a-constraint}
				   \end{align}
\noindent \tb{Step 2}. Any $\P\in \Mp(\X_1 \times \X_2)$ can be parametrized as
				   \begin{align}
					 \P &= \sum_{i,j=1}^2 p_{ij} \delta_{(i,j)}, \quad p_ {ij}\ge 0,\, \forall\, (i,j)\quad\text{and}\quad \sum_{i,j=1}^2p_{ij}=1. \label{Eq:pp}
				   \end{align}
Let $\F = \P - \P_1 \otimes \P_2 = \sum_{i,j=1}^2 a_{ij} \delta_{(i,j)}$; for illustration see Table~\ref{tab:kxl:joint,joint-prod}.
				   \begin{table}
					   \begin{center}
								 \begin{tabular}{@{}c|cc|c@{}}
								  \toprule
								   $\P$: $y\backslash x$& $1$ & $2$ & $\P_2$\\\midrule
								   $1$ & $p_{11}$ & $p_{21}$ & $q_1=p_{11} + p_{21}$\\
								   $2$ & $p_{12}$ & $p_{22}$ & $q_2=p_{12} + p_{22}$ \\\midrule
								   $\P_1$ & $p_1 = p_{11} + p_{12}$ & $p_2 = p_{21} + p_{22}$ \\\bottomrule
								 \end{tabular}\quad $\Rightarrow$
								 \vspace*{0.3cm}
								
								 \begin{tabular}{@{}c|cc@{}}
								  \toprule
								   $\F:=\P-\P_1 \otimes \P_2$& $1$ & $2$\\\midrule
								   $1$ & $a_{11} = p_{11}- (p_{11} + p_{12})(p_{11} + p_{21})$ & $a_{21} = p_{21} - (p_{21} + p_{22})(p_{11} + p_{21})$\\
								   $2$ & $a_{12} = p_{12} - (p_{11} + p_{12})(p_{12} + p_{22})$ & $a_{22} = p_{22} - (p_{21} + p_{22})(p_{12} + p_{22})$\\\bottomrule
								 \end{tabular}
								   \end{center}
					   \caption{Joint ($\P$), joint minus product of the marginals ($\P - \P_1 \otimes \P_2$).} \label{tab:kxl:joint,joint-prod}
					   \end{table}
It follows from step 1 that $\F$ satisfying \eqref{eq:a-constraint} is equivalent to satisfying \eqref{eq:I-char:2a}. Therefore, for the choice of $\F:=\P-\P_1 \otimes \P_2$, we obtain
				   \begin{align}
					 p_{11}- (p_{11} + p_{12})(p_{11} + p_{21}) + p_{22} - (p_{21} + p_{22})(p_{12} + p_{22}) &= 0, \label{eq:kxl:I-char:constraint1}\\
					 p_{12} - (p_{11} + p_{12})(p_{12} + p_{22}) + p_{21} - (p_{21} + p_{22})(p_{11} + p_{21}) &= 0,\label{eq:kxl:I-char:constraint2}
				   \end{align}
where $(p_{ij})_{i,j\in[2]}$ satisfy \eqref{Eq:pp}.
Solving \eqref{Eq:pp}--\eqref{eq:kxl:I-char:constraint2}, we obtain
				   \begin{align*}
					 p_{11} & = \frac{a[1-(a+b)]}{a+b},\,\,\,p_{12}  = \frac{b[1-(a+b)]}{a+b},\,\,\, p_{21} = a\quad\text{and}\quad p_{22}=b,
				   \end{align*}
				   with $0\le a,b \le 1$, $a+b\le 1$ and $(a,b)\ne \b 0$. The resulting distribution family with its marginals is summarized in Table~\ref{tab:eq:kxl:I-char}.
					   \begin{table}
					   \begin{center}
								 \begin{tabular}{@{}c|cc|c@{}}
								  \toprule
								   $\P$: $y\backslash x$& $1$ & $2$ & $\P_2$\\\midrule
								   $1$ & $p_{11}=\frac{a[1-(a+b)]}{a+b}$ & $p_{21}=a$ & $q_1=\frac{a}{a+b}$\\
								   $2$ & $p_{12}=\frac{b[1-(a+b)]}{a+b}$ & $p_{22}=b$ & $q_2=\frac{b}{a+b}$ \\\midrule
								   $\P_1$ & $p_1 = 1-(a+b)$ & $p_2 = a+b$ &\\\bottomrule
								 \end{tabular}
								 \caption{Family of probability distributions solving \eqref{Eq:pp}--\eqref{eq:kxl:I-char:constraint2}. \label{tab:eq:kxl:I-char}}
					   \end{center}
\vspace{-4mm}
					   \end{table}
				   It can be seen that each member of this family (any $a$, $b$ in the constraint set) factorizes: $\P = \P_1 \otimes \P_2$.
				   In other words, $\F=\P - \P_1 \otimes \P_2 = 0$; hence $k_1 \otimes k_2$ is $\I$-characteristic.\vspace{2mm}\\
\textbf{Remark.} We would like to mention that while $k_1$ and $k_2$ are characteristic, they are not universal. Since $\X$ is finite, the usual notion of universality (also called $c$-universality) matches with $c_0$-universality. Therefore, from \eqref{Eq:char-univ}, we have $\int_{\X}\int_{\X} k(x,x')\,\d\F(x)\,\d\F(x)=(a_1-a_2)^2$ where $\F=a_1\delta_1+a_2\delta_2$ for some $a_1,a_2\in\R\backslash\{0\}$. Clearly, the choice of $a_1=a_2$ establishes that there exists $\F\in \Mb(\X)\backslash\{0\}$ such that $\int_{\X}\int_{\X} k(x,x')\,\d\F(x)\,\d\F(x)=0$. Hence $k$ is not universal. Note that the constraint in \eqref{Eq:tt}, which is needed to verify the characteristic property of $k$ is not needed to verify its universality.

\subsection{Proof of Theorem~\ref{thm2:k_m:char<=>xk_m:I-char}}
Define $\H_m:= \H_{k_m}$.\vspace{2mm}\\
$(i)$ Suppose $k_1$ and $k_2$ are characteristic and
		that for some $\F=\P - \P_1 \otimes \P_2\in \I$,
		\begin{align}
		\H_1 \otimes \H_2\ni \int_{\X_1 \times \X_2} \left(k_1 \otimes k_2\right)(\cdot,x)\, \d \F (x) = \int_{\X_1 \times \X_2} k_1(\cdot,x_1) \otimes k_2(\cdot,x_2)\, \d \F (x) = 0, \label{eq:m=0:a}
		\end{align}
		where $x=(x_1,x_2)$. We want to show that $\F = 0$. By the second-countability of $\X_m$-s, the product $\sigma$-field, i.e., $\otimes_{m=1}^2 \B(\X_m)$ generated by the cylinder sets   	
		$B_1 \times \X_2$ and $\X_1 \times B_2$ ($B_m \in \B(\X_m), m=1,2$), coincides with the Borel $\sigma$-field $\B(\X_1 \times \X_2)$ on the product space \cite[Lemma~4.1.7]{dudley04real}:
		\begin{align*}
		    \otimes_{m=1}^2 \B\left(\X_m\right) = \B\left(\X_1 \times \X_2\right).
		\end{align*}
		Hence, it is sufficient to prove that $\F\left(B_1 \times B_2\right) = 0$, $\forall\, B_m \in \B(\X_m)$, $m=1,2$. To this end, it follows from \eqref{eq:m=0:a} that for all $h_2 \in \H_2$,
		\begin{align}
		   \H_1 \ni &\int_{\X_1 \times \X_2} k_1(\cdot,x_1) h_2(x_2)\, \d \F (x) = \int_{\X_1} k_1(\cdot,x_1)\, \d \nu(x_1)=0, \label{eq:I-char:1}
		\end{align}
		where
		\begin{align*}
		  \nu(B_1) &:=\nu_{h_2}(B_1)= \int_{\X_1 \times \X_2} \chi_{B_1}(x_1)h_2(x_2)\, \d \F(x),\quad B_1\in \B(\X_1).
		\end{align*}
		Since $k_1$ is characteristic, \eqref{eq:I-char:1} implies $\nu = 0$, provided that $|\nu|(\X_1)<\infty$ and $\nu(\X_1) = 0$. These two requirements hold:
		\begin{align*}
		  \nu(\X_1) &=\int_{\X_1 \times \X_2}h_2(x_2)\,\d\F(x)=\int_{\X_2} h_2(x_2)\, \d [\P_2-\P_2](x_2) = 0, \nonumber\\
		  |\nu|(\X_1) &\le \int_{\X_1 \times \X_2} \underbrace{\left|h_2(x_2)\right|}_{\big|\left<h_2,k_2(\cdot,x_2)\right>_{\H_2}\big|} \d [\P + \P_1 \otimes \P_2](x_1,x_2) \nonumber\\
&\le \left\| h_2 \right\|_{\H_2} \int_{\X_1\times \X_2} \sqrt{k_2(x_2,x_2)}\, \d [\P + \P_1 \otimes \P_2](x_1,x_2) \nonumber\\
						&\le 2 \left\| h_2 \right\|_{\H_2} \int_{\X_2} \sqrt{k_2(x_2,x_2)}\, \d\P_2(x_2)<\infty, 
		\end{align*}
where the last inequality follows from the boundedness of $k_2$.
		The established $\nu = 0$ implies that for $\forall\, B_1\in \B(\X_1)$ and $\forall\, h_2 \in \H_2$,
		\begin{align*}
		  0 &= \nu(B_1) = \left<h_2,\int_{\X_1 \times \X_2} \chi_{B_1}(x_1)k_2(\cdot,x_2)\, \d \F(x) \right>_{\H_2},\label{eq:I-char:2}
		\end{align*}
		and hence
		\begin{align}
0 = \int_{\X_1 \times \X_2} \chi_{B_1}(x_1)k_2(\cdot,x_2)\, \d \F(x) = \int_{ \X_2} k_2(\cdot,x_2)\, \d \theta_{B_1} (x_2),
		\end{align}
		where
		\begin{align*}
		 \theta_{B_1}(B_2) = \int_{\X_1 \times \X_2} \chi_{B_1}(x_1) \chi_{B_2}(x_2)\, \d \F(x),\quad B_2 \in \B(\X_2).
		\end{align*}
		 Using the characteristic property of $k_2$, it follows from \eqref{eq:I-char:2} that $\theta_{B_1}=0$ for $\forall B_1 \in \B(\X_1)$, i.e.,
		\begin{align*}
		0 &= \theta_{B_1}(B_2) = \F(B_1 \times B_2), \quad \forall\, B_1 \in \B(\X_1),\, \forall\, B_2 \in \B(\X_2)
		\end{align*}
		provided that $\theta_{B_1}(\X_2) = 0$ and $|\theta_{B_1}|(\X_2) <\infty$. Indeed, both these conditions hold:
		\begin{align*}
		   \theta_{B_1}(\X_2) &= \int_{\X_1\times \X_2} \chi_{B_1}(x_1)\,\d\F(x)=\int_{\X_1} \chi_{B_1}(x_1)\, \d [\P_1 - \P_1](x_1) = 0,\\
		   |\theta_{B_1}|(\X_2)&\le \int_{\X_1 \times \X_2}\, \d[\P+\P_1 \otimes \P_2](x) = 2.
		\end{align*}
$(ii)$	Assume w.l.o.g.\ that $k_1$ is not characteristic. This means there exists $\P_1 \ne \P_1' \in \Mp(\X_1)$ such that $\mu_{k_1}(\P_1) = \mu_{k_1}\left(\P_1'\right)$.
		Our goal is to construct an $\F \in \Mp\left(\times_{m=1}^M\X_m\right)$ such that
		\begin{align*}
		  \mu_{\otimes_{m=1}^M k_m}\left(\F - \otimes_{m=1}^M\F_m\right) &= \int_{\times_{m=1}^M} \otimes_{m=1}^M k_m(\cdot,x_m)\, \d \big[\F - \otimes_{m=1}^M \F_m\big] = 0, \text{ but } \F \ne \otimes_{m=1}^M\F_m.
		\end{align*}
Define $\Ib:=\F- \otimes_{m=1}^M\F_m \in\I$. In other words we want to get a witness $\Ib \in \I$ proving that $\otimes_{m=1}^M k_m$ is not $\I$-characteristic. Let us take $z \ne z' \in \X_2$, which is possible since $|\X_2| \ge 2$. Let us define $\F$ as\footnote{The $\F$ construction specializes to that of
		 \citet[Proposition~3.15]{lyons13distance} in the $M=2$ case; Lyons used it for distance covariances, which is known to be equivalent to HSIC \citep{sejdinovic13equivalence}.}
		\begin{align*}
		  \F &= \frac{\P_1 \otimes \delta_{z}\otimes (\otimes_{m=3}^M\Q_m) + \P_1' \otimes \delta_{z'}\otimes (\otimes_{m=3}^M\Q_m)}{2} \in \Mp\left(\times_{m=1}^M \X_m\right).
\end{align*}
It is easy to verify that  $$\F_1 = \frac{\P_1 + \P_1'}{2},\, \F_2 = \frac{\delta_{z} + \delta_{z'}}{2}\,\,\text{and}\,\, \F_m = \Q_m\,\quad (m=3,\ldots,M),$$
where $\Q_3, \ldots, \Q_M$ are arbitrary probability measures on $\X_3, \ldots, \X_M$, respectively. First we check that $\Ib \ne 0$. Indeed it is the case since\vspace{1mm}
		 \begin{compactitem}
		   \item $z \ne z'$ and $\X_2$ is a Hausdorff space, there exists $B_2 \in \B(\X_2)$ such that $z \in B_2$, $z' \not\in B_2$.\vspace{.5mm}
		   \item $\P_1 \ne \P_1'$, $\P_1(B_1) \ne \P_1'(B_1)$ for some $B_1 \in \B(\X_1)$.\vspace{1mm}
		 \end{compactitem}
		 Let $S = B_1 \times B_2 \times \left(\times_{m=3}^M \X_m\right)$, and compare its measure under $\F$ and $\otimes_{m=1}^M \F_m$:
		 \begin{align*}
		   \F(S) &= \frac{\P_1(B_1) \overbrace{\delta_{z}(B_2)}^{=1\, (z \in B_2)}\prod_{m=3}^M \overbrace{\Q_m(\X_m)}^{=1} + \P_1'(B_1) \overbrace{\delta_{z'}(B_2)}^{=0\, (z' \not \in B_2)}\prod_{m=3}^M \overbrace{\Q_m(\X_m)}^{=1}}{2}\\
			  & = \frac{\P_1(B_1)}{2},\\
		   \left(\otimes_{m=1}^M\F_m\right)(S) &= \prod_{m=1}^M\F_m(B_m) = \frac{\P_1(B_1) + \P_1'(B_1)}{2} \frac{\overbrace{\delta_{z}(B_2)}^{=1} + \overbrace{\delta_{z'}(B_2)}^{=0}}{2} \prod_{m=3}^M \overbrace{\Q_m(\X_m)}^{=1}\\
							&= \frac{\P_1(B_1) + \P_1'(B_1)}{4} \ne \frac{\P_1 (B_1)}{2},
		 \end{align*}
		 where the last equality holds since $\P_1(B_1) \ne \P_1'(B_1)$. This shows that $\Ib = \F - \otimes_{m=1}^M\F_m \ne 0$ since $\Ib(S) \ne 0$.

\noindent Next we prove that $\mu_{\otimes_{m=1}^M k_m}\left(\F - \otimes_{m=1}^M\F_m\right)=0$. Indeed,
\begin{eqnarray*}
\mu_{\otimes_{m=1}^M k_m}\left(\Ib \right) &{}={}&
\int_{\times_{m=1}^M\X_m} \otimes_{m=1}^M k_m(\cdot,x_m)\, \d \left[\F - \otimes_{m=1}^M\F_m\right](x_1,\ldots,x_M)\\
			&{}={}& \int_{\times_{m=1}^M\X_m} \otimes_{m=1}^M k_m(\cdot,x_m)\, \d \left(\left[ \frac{\P_1 \otimes \delta_{z} + \P_1' \otimes \delta_{z'}}{2} - \frac{\P_1 + \P_1'}{2} \otimes \frac{\delta_{z} + \delta_{z'}}{2}\right] \right. \\
			 &{}{}&\quad\quad\quad\quad\quad\quad \left. \otimes \left(\otimes_{m=3}^M\Q_m\right) \right) (x_1,\ldots,x_M)\\
			&{}={}& \int_{\times_{m=1}^M\X_m} \otimes_{m=1}^M k_m(\cdot,x_m)\, \d \left( \left[ \frac{\P_1(x_1) \otimes \delta_z(x_2) + \P_1'(x_1) \otimes \delta_{z'}(x_2)}{2} \right. \right. \\
				&{}{}&\quad\quad\quad\quad\left. - \frac{\P_1(x_1) \otimes \delta_z(x_2) + \P_1(x_1) \otimes \delta_{z'}(x_2)}{4}\right.\\
&{}{}&\quad\quad\quad\quad\quad\quad\left.\left. - \frac{\P_1'(x_1) \otimes \delta_z(x_2) + \P_1'(x_1) \otimes \delta_{z'}(x_2)}{4} \right]
				 \otimes(\otimes_{m=3}^M\Q_m(x_m))\right)\\
				&{}\stackrel{(*)}{=}{}& \left[ \frac{\mu_{k_1}(\P_1) \otimes k_2(\cdot,z) + \mu_{k_1}(\P_1') \otimes k_2(\cdot,z')}{2}\right.\\
&{}{}&\quad\quad- \frac{\mu_{k_1}(\P_1) \otimes k_2(\cdot,z) + \mu_{k_1}(\P_1) \otimes k_2(\cdot,z')}{4} \\
				&{}{}&\quad\quad\quad\quad\left.  - \frac{\mu_{k_1}(\P_1') \otimes k_2(\cdot,z) + \mu_{k_1}(\P_1') \otimes k_2(\cdot,z')}{4}\right] \otimes \left[\otimes_{m=3}^M \mu_{k_m}\left(\Q_m\right)\right]\\
				&{}={}& \underbrace{0}_{\in\, \H_{k_1\otimes k_2}} \otimes \left[ \otimes_{m=3}^M \mu_{k_m}\left(\Q_m\right) \right] = 0,
		 \end{eqnarray*}
		 where we used $\mu_{k_1}(\P_1) = \mu_{k_1}\left(\P_1'\right)$ in $(*)$.

\subsection{Proof of Example~\ref{example:k_m:char=does not=>x_m k_m:I-char}}\label{subsec:ex2}
Let $M=3$, $\times_{m=1}^M\X_m = \{(i_1,i_2,i_3): i_m\in\{1,2\},\,m\in[3]\}$, $k_m(x,x') = 2\delta_{x,x'} - 1$. Our goal is to show that $\otimes_{m=1}^3 k_m$ is \emph{not} $\I$-characteristic. The structure of the proof is as follows:\vspace{1mm}
 \begin{compactenum}
   \item First we describe the equations of the non-characteristic property of $\otimes_{m=1}^3 k_m$ with a general finite signed measure $\F= \sum_{i_1,i_2,i_3=1}^2 a_{i_1,i_2,i_3} \delta_{(i_1,i_2,i_3)}$
on $\times_{m=1}^3\X_m$ where $a_{i_1,i_2,i_3}\in \R$ ($\forall\, i_1, i_2, i_3$).\vspace{.5mm}
   \item Next, we apply the $\F = \P - \otimes_{m=1}^3 \P_m$ parameterization and show that there exists $\P$ that satisfies the equations of step 1 to conclude that $\otimes_{m=1}^3 k_m$ is not $\I$-characteristic.\vspace{2mm}
 \end{compactenum}
\noindent The details are as follows.\vspace{2mm}\\
\noindent \tb{Step 1}. The equations of non-characteristic property in terms of $\b A = \left[ a_{i_1,i_2,i_3} \right]_{(i_m)^3_{m=1}\in[2]^3}\in \R^{2\times 2 \times 2}$ are
		\begin{align}
			 \F &\in \Mb\left(\times_{m=1}^3\X_m\right)\backslash\{0\} \Leftrightarrow \b A\ne \b 0,\nonumber\\
			 0&=\F(\times_{m=1}^3\X_m)  \Leftrightarrow 0 = \sum_{i_1,i_2,i_3=1}^2 a_{i_1,i_2,i_3} \label{eq:M3:eq1},\\
			 0 &= \int_{\times_{m=1}^3\X_m}\int_{\times_{m=1}^3\X_m} \underbrace{(\otimes_{m=1}^3k_m)\left((i_1,i_2,i_3),(i_1',i_2',i_3')\right)}_{\prod_{m=1}^3 k_m(i_m,i_m')}\,\d \F(i_1,i_2,i_3)\, \d \F(i_1',i_2',i_3') \nonumber\\
			 &= \sum_{i_1,i_2,i_3=1}^2 \sum_{i_1',i_2',i_3'=1}^2 \prod_{m=1}^3 k_m(i_m,i_m')a_{i_1,i_2,i_3}a_{i_1',i_2',i_3'}. \label{eq:M3:eq2}
		 \end{align}
	Solving \eqref{eq:M3:eq1} and \eqref{eq:M3:eq2}
yields
	\begin{align*}
	 a_{1,1,1} + a_{1,2,2} + a_{2,1,2} + a_{2,2,1} &= 0 \quad \text{and}\quad  a_{1,1,2} + a_{1,2,1} + a_{2,1,1} + a_{2,2,2} =0.
	\end{align*}
\noindent \tb{Step 2}. The equations of non $\I$-characteristic property can be obtained from step 1 by choosing $\F = \P - \otimes_{m=1}^M \P_m$, where
    $$   \P= \sum_{i_1,i_2,i_3=1}^2 p_{i_1,i_2,i_3} \delta_{(i_1,i_2,i_3)} \quad\text{and}\quad \b P= \left[p_{i_1,i_2,i_3}\right]_{(i_m)^3_{m=1}\in[2]^3}\in \R^{2\times 2 \times 2}.$$
In other words, it is sufficient to obtain a $\b P$ that solves the
   following system of equations for which $\b A=\b A(\b P)\ne \b 0$:
	\begin{align}
	 \sum_{i_1,i_2,i_3=1}^2 p_{i_1,i_2,i_3} &= 1,\label{eq:z:first}\\
	 p_{i_1,i_2,i_3}& \ge 0, \, \forall\,(i_1,i_2,i_3)\in[2]^3,\\
	 a_{1,1,1} + a_{1,2,2} + a_{2,1,2} + a_{2,2,1} &= 0, \label{ex2:a-eq:1}\\
	 a_{1,1,2} + a_{1,2,1} + a_{2,1,1} + a_{2,2,2} &=0,\label{ex2:a-eq:2}
	 \end{align}
	 where
\begin{equation}
	  a_{i_1,i_2,i_3} = p_{i_1,i_2,i_3} - p_{1,i_1} p_{2,i_2} p_{3,i_3},
\end{equation}
and
\begin{equation}
	  p_{1,i_1} = \sum_{i_2,i_3=1}^2 p_{i_1,i_2,i_3}, \quad
	  p_{2,i_2} = \sum_{i_1,i_3=1}^2 p_{i_1,i_2,i_3}, \quad
	  p_{3,i_3} = \sum_{i_1,i_2=1}^2 p_{i_1,i_2,i_3}.\label{eq:z:last}
	\end{equation}
	One can get an analytical description for the solution of \eqref{eq:z:first}--\eqref{eq:z:last}, where the solution $\b P(\b{z})$ is parameterized by $\b z=(z_0,\ldots,z_5)\in\R^6$. For explicit expressions, we refer the reader to Appendix~\ref{app:z-nasty}. In the following, we present two examples of $\b P$ that satisfy \eqref{eq:z:first}--\eqref{eq:z:last} such that $\b A\ne \b 0$, thereby establishing the non $\I$-characteristic property of $\otimes^3_{m=1}k_m$. \vspace{1mm}
	 \begin{compactenum}
	   \item
$\b P$:
		   \begin{align*}
			 p_{1,1,1} &= \frac{1}{5}, & p_{1,1,2} &= \frac{1}{10}, & p_{1,2,1} &= \frac{1}{10}, & p_{1,2,2} &= \frac{1}{10},\\
			 p_{2,1,1} &= \frac{1}{5}, & p_{2,1,2} &= \frac{1}{10}, & p_{2,2,1} &= \frac{1}{10}, & p_{2,2,2} &= \frac{1}{10},
		   \end{align*}
		  and $\b A$:
		   \begin{align}
			 a_{1,1,1} &= \frac{1}{50},& a_{1,1,2} &= -\frac{1}{50}, & a_{1,2,1} &= -\frac{1}{50}, & a_{1,2,2} &= \frac{1}{50}, \label{eq:I-char:A1:row1}\\
			 a_{2,1,1} &= \frac{1}{50}, & a_{2,1,2} &= -\frac{1}{50}, & a_{2,2,1} &= -\frac{1}{50}, & a_{2,2,2} &= \frac{1}{50}.\label{eq:I-char:A1:row2}
		   \end{align}
	   \item
$\b P$:
		   \begin{align*}
			 p_{1,1,1} &= 0, & p_{1,1,2} &= \frac{1}{10}, & p_{1,2,1} &= \frac{1}{10}, & p_{1,2,2} &= \frac{1}{10},\\
			 p_{2,1,1} &= \frac{1}{10}, & p_{2,1,2} &= \frac{1}{10}, & p_{2,2,1} &= \frac{3}{10}, & p_{2,2,2} &= \frac{1}{5},
		   \end{align*}
		  and $\b A$:
		   \begin{align*}
			 a_{1,1,1} &= -\frac{9}{200},& a_{1,1,2} &= \frac{11}{200}, & a_{1,2,1} &= -\frac{1}{200}, & a_{1,2,2} &= -\frac{1}{200},\\
			 a_{2,1,1} &= -\frac{1}{200}, & a_{2,1,2} &= -\frac{1}{200}, & a_{2,2,1} &= \frac{11}{200}, & a_{2,2,2} &= -\frac{9}{200}.
		   \end{align*}
	 \end{compactenum}
In fact these examples are obtained with the choices $\b z=(\frac{1}{10}, \frac{1}{10},\frac{1}{10},\frac{1}{10},\frac{1}{10},\frac{1}{10})$
and $\b z=(\frac{3}{10}, \frac{1}{10},\frac{1}{10},\frac{1}{10},\frac{1}{10},\frac{2}{10})$ respectively. See Appendix~\ref{app:z-nasty} for details.

\subsection{Proof of Theorem~\ref{thm4:contboundedshiftinv}}
It follows from \eqref{eq:char-relations} and Remark~\ref{Rem:remark}(iii) that $(v)\Rightarrow (iii) \Rightarrow (ii) \Leftrightarrow (i)$. 
It also follows from \eqref{eq:char-relations} and Theorem~\ref{thm2:k_m:char<=>xk_m:I-char}(ii) that
$(v)\Rightarrow (iv)\Rightarrow (i)$. We now show that $(i)\Rightarrow (v)$ which establishes the equivalence of $(i)$--$(v)$. Suppose $(i)$ holds. Then by Bochner's theorem \citep[Theorem 6.6]{W05}, we have that for all $m\in [M]$,
			\begin{align*}
			  k_m(x_m,y_m) = \int_{\R^{d_m}} e^{-\sqrt{-1}\langle \omega_m, x_m- y_m\rangle}\, \d \Lambda_m(\omega_m),\,\,x_m,y_m\in\R^{d_m},
			\end{align*}
where $(\Lambda_m)^M_{m=1}$ are finite non-negative Borel measures on $(\R^{d_m})^M_{m=1}$ respectively. This implies $$\otimes^M_{m=1}k_m(x_m,y_m)=\otimes^M_{m=1} \int_{\R^{d_m}} e^{-\sqrt{-1}\langle \omega_m, x_m- y_m\rangle}\, \d \Lambda_m(\omega_m)=\int_{\R^d}e^{-\sqrt{-1}\langle \omega,x-y\rangle}\,\d\Lambda(\omega),$$
where $x=(x_1,\ldots,x_M)\in\R^d$, $y=(y_1,\ldots,y_M)\in\R^d$, $\omega=(\omega_1,\ldots,\omega_M)\in\R^d$, $d=\sum^M_{m=1}d_m$ and $\Lambda:=\otimes^M_{m=1}\Lambda_m$. \citet[Theorem~9]{sriperumbudur10hilbert} showed that $k_m$ is characteristic iff $\text{supp}\left(\Lambda_m\right) = \R^{d_m}$, where $\text{supp}(\cdot)$ denotes the support of its argument. Since $\text{supp}(\Lambda)=\text{supp}\left(\otimes^M_{m=1}\Lambda_m\right) = \times_{m=1}^M \text{supp}\left(\Lambda_{m}\right)=\times_{m=1}^M \R^{d_m}=\R^d$, it follows that $\otimes^M_{m=1}k_m$ is characteristic.

\subsection{Proof of Theorem~\ref{thm4:prod-of-c0universal}} \label{sec:proof:c_0-univ}
  The $c_0$-kernel property of $k_m$-s ($m=1,\ldots,M$) implies that of $\otimes_{m=1}^M k_m$. Moreover, $\X_m$-s are LCP spaces, hence $\times_{m=1}^M \X_m$ is also LCP. \vspace{2mm}\\
($\Leftarrow$) Assume that $\otimes_{m=1}^Mk_m$ is $c_0$-universal. Since $\otimes_{m=1}^M \Mb\left(\X_m\right) \subseteq \Mb\left(\times_{m=1}^M\X_m\right)$, we have that for all $\F = \otimes_{m=1}^M \F_m \in \otimes_{m=1}^M\Mb(\X_m) \backslash \{0\}$,
        \begin{align*}
            0 &< \int_{\times_{m=1}^M \X_m} \int_{\times_{m=1}^M \X_m} \underbrace{\left(\otimes_{m=1}^Mk_m\right)(x,x')}_{\prod_{m=1}^M k_m(x_m,x_m')}\, \d \F(x)\, \d \F(x')\\
            &= \prod_{m=1}^M \int_{\X_m \times \X_m} k_m(x_m,x_m')\, \d \F_m(x_m)\,\d \F_m\left(x_m'\right),
        \end{align*}
        where $x=(x_1,\ldots,x_M)$ and $x' = \left(x_1',\ldots,x_M'\right)$. The above inequality implies $$\int_{\X_m \times \X_m} k_m(x_m,x_m')\, \d \F_m(x_m)\,\d \F_m\left(x_m'\right)>0,\,\,\forall\,m\in[M].$$ Since $\F\in \otimes^M_{m=1}\Mb\left(\X_m\right) \backslash \{0\}$ iff $\F_m \in \Mb(\X_m) \backslash \{0\}$ for all $m\in[M]$, the result follows. \vspace{2mm}\\
($\Rightarrow$) Assume that $k_m$-s are $c_0$-universal. By the note above $\otimes_{m=1}^Mk_m$ is $c_0$-kernel; its $c_0$-universality is equivalent to the injectivity of $\mu = \mu_{\otimes_{m=1}^Mk_m}$ on $\Mb\left(\times_{m=1}^M\X_m\right)$. In other words, we want to prove that $\mu(\F) = 0$ implies $\F=0$, where $\F \in \Mb\left(\times_{m=1}^M\X_m\right)$. We will use the shorthand $\H_m = \H_{k_m}$ below.

Suppose there exists $\F \in \Mb\left(\times_{m=1}^M\X_m\right)$ such that
		  \begin{align}
			\mu_{\F} = \int_{\times_{m=1}^M\X_m} \underbrace{\left(\otimes_{m=1}^M k_m\right)(\cdot,x)}_{\otimes_{m=1}^M k_m(\cdot,x_m)}\, \d \F (x)= 0\hspace{0.1cm} (\in \otimes_{m=1}^M\H_m). \label{eq:m=0}
		  \end{align}		
		  Since $\X_m$-s are LCP, $\otimes_{m=1}^M \B\left(\X_m\right) = \B\left(\times_{m=1}^M\X_m\right)$ \cite[page~480]{steinwart08support}. Hence, in order to get $\F = 0$ it is sufficient to prove that
		  \begin{align*}
			\F\left(\times_{m=1}^M B_m\right) = 0, \quad \forall\, B_m \in \B(\X_m), m\in[M].
		  \end{align*}
		   We will prove by induction that for $m=0,\ldots,M$
		  \begin{align}
		  \left(\otimes_{j=m+1}^M \H_j\ni\right) 0 &= \int_{\times_{j=1}^M \X_j} \prod_{j=1}^m \chi_{B_j}(x_j) \otimes_{j=m+1}^M k_j(\cdot,x_j)\,\d \F(x) \nonumber\\
		  &=: o(B_1,\ldots,B_m,k_{m+1},\ldots,k_M), \forall\, B_j \in \B(\X_j),\, j\in[m],\label{induction:m}
		  \end{align}
		  which
\begin{itemize}
\item[($*$)] reduces to \eqref{eq:m=0} when $m=0$ by defining $\prod_{j=1}^0 \chi_{B_j}(x_j):=1$;\vspace{-1.5mm}
\item[($\dagger$)] for $m=M$, $\otimes_{m=M+1}^M \H_m$ is defined to be equal to $\R$ and $\otimes_{m=M+1}^M k_m(\cdot,x_m):=1$, in which case $o(B_1,\ldots,B_M) = \F\left(\times_{j=1}^M B_j\right)=0\Rightarrow \F=0$, the result we want to prove.
\end{itemize}
From the above, it is clear that \eqref{induction:m} holds for $m=0$. Assuming \eqref{induction:m} holds for some $m$, we now prove that it holds for $m+1$. To this end, it follows from \eqref{induction:m} that $\forall\, h_{m+2}\in \H_{m+2}, \ldots, \forall\, h_M \in \H_M$,
		  \begin{align*}
		  (\H_{m+1} \ni)\,0 &= o(B_1,\ldots,B_m,k_{m+1},\ldots,k_M) \left(h_{m+2},\ldots,h_{M}\right)\\
			   &=\left[\int_{\times_{j=1}^M \X_j} \left( \prod_{j=1}^m \chi_{B_j}(x_j) \right) \otimes_{j=m+1}^M k_j(\cdot,x_j)\, \d \F(x)\right] (h_{m+2},\ldots,h_{M})\\
			   &= \int_{\times_{j=1}^M \X_j} k_{m+1}(\cdot,x_{m+1}) \prod_{j=1}^m \chi_{B_j}(x_j) \prod_{j=m+2}^Mh_j(x_j)\, \d \F(x) \\
			   &= \int_{\X_{m+1}} k_{m+1}(\cdot,x_{m+1})\, \d \nu(x_{m+1}),
		  \end{align*}
		  where
		  \begin{align*}
			\nu(B)&:=\nu_{B_1,\ldots,B_m,h_{m+2},\ldots,h_{M}}(B)\\
			   &= \int_{\times_{j=1}^M \X_j} \left[\prod_{j=1}^m \chi_{B_j}(x_j)\right] \chi_B(x_{m+1}) \left[\prod_{j=m+2}^M h_j(x_j)\right] \d \F(x),\, B \in \B(\X_{m+1}).
		  \end{align*}
		  By the $c_0$-universality of $k_{m+1}$,
		  \begin{align}
			\nu = 0 \text{ for } \forall\, h_{m+2}\in\H_{m+2},\ldots,\forall\, h_{M}\in \H_M \label{eq:X-measure-null}
		  \end{align}
		  provided that $\nu \in \Mb(\X_{m+1})$, in other words if $|\nu|(\X_{m+1}) < \infty$. This condition is met:
				\begin{align*}
				  |\nu|(\X_{m+1}) &\le \int_{\times_{j=1}^M \X_j} \prod_{j=m+2}^M\underbrace{\left|\left<h_j,k_j(\cdot,x_j)\right>_{\H_j}\right|}_{\le \left\|h_j\right\|_{\H_j}\sqrt{k_j(x_j,x_j)}} \d |\F|(x)\nonumber\\
				  &\le |\F|\left(\times_{m=1}^M\X_m\right) \prod_{j=m+2}^M \left\|h_j\right\|_{\H_j} \sup_{x\in \X_j, x'\in \X_j} \sqrt{k_j(x,x')}<\infty,
				\end{align*}
where we used the boundedness of $k_m$-s in the last inequality. \eqref{eq:X-measure-null} implies that for $\forall\, B_1\in \B(\X_1),\ldots, \forall\, B_{m+1}\in \B(\X_{m+1})$ and $\forall\, h_{m+2}\in \H_{m+2},\ldots, \forall\, h_{M}\in \H_M$
		  \begin{align*}
			0 &= \nu(B_{m+1}) = \int_{\times_{j=1}^M\X_j} \left[ \prod_{j=1}^{m+1}\chi_{B_j}(x_j)\right] \left[\prod_{j=m+2}^M h_j(x_j)\right]\, \d \F(x)\\
			 &= \left<\otimes_{j=m+2}^M h_j, \int_{\times_{j=1}^{M}\X_j} \left[\prod_{j=1}^{m+1} \chi_{B_j}(x_j) \right] \otimes_{j=m+2}^M k_j(\cdot,x_j)\,\d \F(x) \right>_{\otimes_{j=m+2}^M\H_j},
		  \end{align*}
		  and therefore
		  \begin{align*}
			o(B_1,\ldots,B_{m+1},k_{m+2},\ldots,k_M) &= \int_{\times_{j=1}^{M}\X_j} \left[\prod_{j=1}^{m+1} \chi_{B_j}(x_j) \right] \otimes_{j=m+2}^M k(\cdot,x_j)\,\d \F(x)\\
			& = 0 \left(\in \otimes_{j=m+2}^M\H_j\right)
		  \end{align*}
		  for $\forall\, B_1\in\B(\X_1), \ldots,\forall\, B_{m+1}\in \B(\X_{m+1})$, i.e., \eqref{induction:m} holds for $m+1$. Therefore, by induction, \eqref{induction:m} holds for $m=M$ and the result follows from ($\dagger$). To justify the convention in ($\dagger$), consider the case of $m=M-1$ in which case \eqref{induction:m} can be written as $$\int_{\X_M} k_M(\cdot,x_M)\, \d \nu(x_M)=0,$$ where
$$\nu(B) = \int_{\times_{j=1}^M\X_j} \left[\prod_{j=1}^{M-1}\chi_{B_j}(x_j)\right] \chi_B(x_M)\, \d \F(x), \, B\in \B(\X_M).$$
Then by the $c_0$-universal property of $k_M$, since
		  \begin{align*}
			|\nu|(\X_M) & \le \int_{\times_{j=1}^M\X_j} 1\, \d |\F|(x) = |\F|\left(\times_{j=1}^M\X_j\right) <\infty
		  \end{align*}
we obtain
		  \begin{align*}
			\int_{\times_{j=1}^M \X_j} \prod_{j=1}^M \chi_{B_j}(x_j)\, \d \F(x) = \F\left(\times_{j=1}^M B_j\right) = 0, \forall\, B_1 \in \B(\X_1), \ldots, \forall\, B_M \in \B(\X_M).
		  \end{align*}

\subsection{Proof of Example~\ref{example:k1,k2:univ,k_3:char=does not=>x_m k_m:I-char}}\label{subsec:ex3}
The proof follows by a simple modification of that of Example~\ref{example:k_m:char=does not=>x_m k_m:I-char} (Section~\ref{subsec:ex2}). 
The equations of a witness $\b A=[a_{i_1,i_2,i_3}]_{(i_m)_{m=1}^3\in [2]^3}$ (and corresponding $\b P=[p_{i_1,i_2,i_3}]_{(i_m)_{m=1}^3\in [2]^3}$)
for the non-$\I$-characteristic property of $\otimes_{m=1}^3k_m$ take the form:
		\begin{align}
			 \b A &\ne \b 0,\nonumber\\
			 0 &= \sum_{i_1,i_2,i_3=1}^2 a_{i_1,i_2,i_3},\label{ex3:constraint-2} \\
			 0 &= \sum_{i_1,i_2,i_3=1}^2 \sum_{i_1',i_2',i_3'=1}^2 \prod_{m=1}^3 k_m(i_m,i_m')a_{i_1,i_2,i_3}a_{i_1',i_2',i_3'}  \nonumber\\
			  &= \left(a_{1,1,1} - a_{2,1,1}\right)^2 + \left(a_{1,1,2} - a_{2,1,2}\right)^2  + \left(a_{1,2,1} - a_{2,2,1}\right)^2  + \left(a_{1,2,2} - a_{2,2,2}\right)^2, \label{ex3:constraint-3}
		 \end{align}
		 where \eqref{ex3:constraint-2} and \eqref{ex3:constraint-3} are equivalent to
		 \begin{align}
		      0 & = \sum_{i_1,i_2,i_3=1}^2 a_{i_1,i_2,i_3}, &
		      a_{1,1,1} & = a_{2,1,1}, &  a_{1,1,2} &= a_{2,1,2}, & a_{1,2,1} &= a_{2,2,1}, & a_{1,2,2} &= a_{2,2,2}. \label{eq:ex3:A}
		 \end{align}
		 While \eqref{eq:ex3:A} is more restrictive than \eqref{ex2:a-eq:1} and \eqref{ex2:a-eq:2} (hence its solution set might even be empty), 
		 one can immediately see that the example of $\b A\ne \b 0$ given in \eqref{eq:I-char:A1:row1} and \eqref{eq:I-char:A1:row2} fulfills 
		 \eqref{eq:ex3:A} proving the non-$\I$-characteristic property of $\otimes_{m=1}^3k_m$.

\newpage
% Acknowledgements should go at the end, before appendices and references
\acks{The authors profusely thank Ingo Steinwart for fascinating discussions on topics related to the paper and for contributing to Remark~\ref{Rem:new}. The authors also thank the anonymous reviewers for their constructive comments that improved the manuscript. A part of the work was carried out while BKS was visiting ZSz at 
CMAP, \'{E}cole Polytechnique. BKS is supported by NSF-DMS-1713011 and also thanks CMAP and DSI for their generous support. ZSz is highly grateful for the Greek hospitality around the Aegean Sea; it greatly contributed to the development of the induction arguments.}

\appendix

\section{Analytical Solution to \eqref{eq:z:first}--\eqref{eq:z:last} in Example~\ref{example:k_m:char=does not=>x_m k_m:I-char}}\label{app:z-nasty}
	The solution of \eqref{eq:z:first}--\eqref{eq:z:last} takes the form
	\begin{align*}
	   p_{1,1,1}& = - \frac{\begin{aligned}
				&z_2 + z_1 + z_4 + z_5 - 3 z_2 z_1 - 4 z_2 z_4 - 4 z_1 z_4 - z_2 z_3 - 2 z_2 z_0 - 2 z_1 z_3 - 3 z_2 z_5\\
				&- 2 z_4 z_3 - z_1 z_0 - 3 z_1 z_5 - 2 z_4 z_0 - 4 z_4 z_5 - z_3 z_0 - z_3 z_5 - z_0 z_5 + 2 z_2 z_1^2 + 2 z_2^2 z_1\\
				& + 4 z_2 z_4^2 + 2 z_2^2 z_4 + 4 z_1 z_4^2 + 2 z_1^2 z_4 + 2 z_2^2 z_0 + 2 z_1^2 z_3 + 2 z_2 z_5^2 + 2 z_2^2 z_5 + 2 z_4^2 z_3\\
				& + 2 z_1 z_5^2 + 2 z_1^2 z_5 + 2 z_4^2 z_0 + 2 z_4 z_5^2 + 4 z_4^2 z_5 - z_2^2 - z_1^2 - 3 z_4^2 + 2 z_4^3 - z_5^2\\
				& + 6 z_2 z_1 z_4 + 2 z_2 z_1 z_3 + 2 z_2 z_4 z_3 + 2 z_2 z_1 z_0 + 4 z_2 z_1 z_5 + 4 z_2 z_4 z_0 + 4 z_1 z_4 z_3\\
				&+ 6 z_2 z_4 z_5 + 2 z_1 z_4 z_0 + 6 z_1 z_4 z_5 + 2 z_2 z_3 z_0 + 2 z_2 z_3 z_5 + 2 z_1 z_3 z_0 + 2 z_2 z_0 z_5\\
				&+ 2 z_1 z_3 z_5 + 2 z_4 z_3 z_0 + 2 z_4 z_3 z_5 + 2 z_1 z_0 z_5 + 2 z_4 z_0 z_5
	              \end{aligned}}
				 {\begin{aligned}
				 &2 z_2 z_1 - z_1 - 2 z_4 - z_3 - z_0 - 2 z_5 - z_2 + 2 z_2 z_4 + 2 z_1 z_4 + 2 z_2 z_0 + 2 z_1 z_3 + 2 z_2 z_5\\
				 &+ 2 z_4 z_3 + 2 z_1 z_5 + 2 z_4 z_0 +
				  4 z_4 z_5 + 2 z_3 z_0 + 2 z_3 z_5 + 2 z_0 z_5 + 2 z_4^2 + 2 z_5^2
				  \end{aligned}},\\
	   p_{1,1,2} &= z_2,\\
	   p_{1,2,1} &= z_1,\\
	   p_{1,2,2} &= z_4,\\
	   p_{2,1,1} &= -
	  \frac{\begin{aligned}
		  & z_4 + z_3 + z_0 + z_5 - z_2 z_1 - z_2 z_4 - z_1 z_4 - z_2 z_3 - 2 z_2 z_0 - 2 z_1 z_3 - 2 z_2 z_5\\
		  &- 3 z_4 z_3 - z_1 z_0 - 2 z_1 z_5 - 3 z_4 z_0 - 4 z_4 z_5 - 3 z_3 z_0 - 4 z_3 z_5 - 4 z_0 z_5 + 2 z_2 z_0^2\\
		  &+ 2 z_1 z_3^2 + 2 z_2 z_5^2 + 2 z_4 z_3^2 + 2 z_4^2 z_3 + 2 z_1 z_5^2 + 2 z_4 z_0^2 + 2 z_4^2 z_0 + 4 z_4 z_5^2 + 2 z_4^2 z_5\\
		  &+ 2 z_3 z_0^2 + 2 z_3^2 z_0 + 4 z_3 z_5^2 + 2 z_3^2 z_5 + 4 z_0 z_5^2 + 2 z_0^2 z_5 - z_4^2 - z_3^2 - z_0^2 - 3 z_5^2\\
		  &+ 2 z_5^3 + 2 z_2 z_1 z_3 + 2 z_2 z_4 z_3 + 2 z_2 z_1 z_0 + 2 z_2 z_1 z_5 + 2 z_2 z_4 z_0 + 2 z_1 z_4 z_3\\
		  &+ 2 z_2 z_4 z_5 + 2 z_1 z_4 z_0 + 2 z_1 z_4 z_5 + 2 z_2 z_3 z_0 + 2 z_2 z_3 z_5 + 2 z_1 z_3 z_0 + 4 z_2 z_0 z_5\\
		  &+ 4 z_1 z_3 z_5 + 4 z_4 z_3 z_0 + 6 z_4 z_3 z_5 + 2 z_1 z_0 z_5 + 6 z_4 z_0 z_5 + 6 z_3 z_0 z_5\\
		 \end{aligned}}%
		 {\begin{aligned}
		  & 2 z_2 z_1 - z_1 - 2 z_4 - z_3 - z_0 - 2 z_5 - z_2 + 2 z_2 z_4 + 2 z_1 z_4 + 2 z_2 z_0 + 2 z_1 z_3+ 2 z_2 z_5 \\
		  &+ 2 z_4 z_3 + 2 z_1 z_5 + 2 z_4 z_0 + 4 z_4 z_5 + 2 z_3 z_0 + 2 z_3 z_5 + 2 z_0 z_5 + 2 z_4^2 + 2 z_5^2\\
		  \end{aligned}},\\
	   p_{2,1,2} &= z_3,\\
	   p_{2,2,1} &= z_0,\\
	   p_{2,2,2} &= z_5,
	\end{align*}
	form, where $\b z=(z_0,z_1,\ldots,z_5)\in\R^6$ satisfies
	\begin{align*}
	   0 &\le \left(2 z_0 z_2 - z_1 - z_2 - z_3 - 2 z_4 - 2 z_5 - z_0 + 2 z_0 z_3 + 2 z_1 z_2 + 2 z_0 z_4 + 2 z_1 z_3 + 2 z_0 z_5\right.\\
	   & \quad \left. + 2 z_1 z_4 + 2 z_1 z_5 + 2 z_2 z_4 + 2 z_2 z_5 + 2 z_3 z_4 + 2 z_3 z_5 + 4 z_4 z_5 + 2 z_4^2 + 2 z_5^2 \right) \times\\
	   & \quad \left(z_0 z_3 - z_3 - z_4 - z_5 - z_0 z_1 - z_0 - z_1 z_2 + z_0 z_5 - 2 z_1 z_4 - z_2 z_3 - z_1 z_5 - 2 z_2 z_4 - z_2 z_5\right.\\
	   & \quad \left. + z_3 z_5 + 2 z_0 z_2^2 + 2 z_1 z_2^2 + 2 z_1^2 z_2 + 2 z_0 z_4^2 + 2 z_1^2 z_3 + 4 z_1 z_4^2 + 2 z_1^2 z_4 + 2 z_1 z_5^2 + 4 z_2 z_4^2 \right.\\
	   & \quad \left. + 2 z_1^2 z_5 + 2 z_2^2 z_4 + 2 z_2 z_5^2 + 2 z_3 z_4^2 + 2 z_2^2 z_5 + 2 z_4 z_5^2 + 4 z_4^2 z_5 - z_1^2 - z_2^2 - z_4^2 + 2 z_4^3 + z_5^2 \right.\\
	   & \quad \left. + 2 z_0 z_1 z_2 + 2 z_0 z_1 z_3 + 2 z_0 z_1 z_4 + 2 z_0 z_2 z_3 + 2 z_0 z_1 z_5 + 4 z_0 z_2 z_4 + 2 z_1 z_2 z_3 + 2 z_0 z_2 z_5\right.\\
	   & \quad \left. + 2 z_0 z_3 z_4 + 6 z_1 z_2 z_4 + 4 z_1 z_2 z_5 + 4 z_1 z_3 z_4 + 2 z_0 z_4 z_5 + 2 z_1 z_3 z_5 +  2 z_2 z_3 z_4 + 6 z_1 z_4 z_5\right.\\
	   & \quad \left. + 2 z_2 z_3 z_5 + 6 z_2 z_4 z_5 + 2 z_3 z_4 z_5 \right),\\\\
	   0 &\le \left(2 z_0 z_2 - z_1 - z_2 - z_3 - 2 z_4 - 2 z_5 - z_0 + 2 z_0 z_3 + 2 z_1 z_2 + 2 z_0 z_4 + 2 z_1 z_3 + 2 z_0 z_5\right. \\
	    & \quad \left.+ 2 z_1 z_4 + 2 z_1 z_5 + 2 z_2 z_4 + 2 z_2 z_5 + 2 z_3 z_4 + 2 z_3 z_5 + 4 z_4 z_5 + 2 z_4^2 + 2 z_5^2\right)  \times \\
	   &\quad \left( z_1 z_2 - z_2 - z_4 - z_5 - z_0 z_1 - z_0 z_3 - z_1 - z_0 z_4 - 2 z_0 z_5 + z_1 z_4 - z_2 z_3 + z_2 z_4\right.\\
	   &\quad \left. - z_3 z_4 - 2 z_3 z_5 + 2 z_0^2 z_2 + 2 z_0 z_3^2 + 2 z_0^2 z_3 + 2 z_0 z_4^2 + 2 z_1 z_3^2 + 2 z_0^2 z_4 + 4 z_0 z_5^2 + 2 z_0^2 z_5\right.\\
	   &\quad \left. + 2 z_1 z_5^2 + 2 z_2 z_5^2 + 2 z_3 z_4^2 + 2 z_3^2 z_4 + 4 z_3 z_5^2 + 2 z_3^2 z_5 + 4 z_4 z_5^2 + 2 z_4^2 z_5 - z_0^2 - z_3^2 + z_4^2\right.\\
	   &\quad \left. - z_5^2 + 2 z_5^3 + 2 z_0 z_1 z_2 + 2 z_0 z_1 z_3 + 2 z_0 z_1 z_4 + 2 z_0 z_2 z_3 + 2 z_0 z_1 z_5 + 2 z_0 z_2 z_4 + 2 z_1 z_2 z_3 \right.\\
	   &\quad \left. + 4 z_0 z_2 z_5 + 4 z_0 z_3 z_4 + 6 z_0 z_3 z_5 + 2 z_1 z_2 z_5 + 2 z_1 z_3 z_4 + 6 z_0 z_4 z_5 + 4 z_1 z_3 z_5 + 2 z_2 z_3 z_4\right.\\
	   &\quad \left.+ 2 z_1 z_4 z_5 + 2 z_2 z_3 z_5 + 2 z_2 z_4 z_5 + 6 z_3 z_4 z_5\right),\\\\
	   & 2 z_0 z_2 + 2 z_0 z_3 + 2 z_1 z_2 + 2 z_0 z_4 + 2 z_1 z_3 + 2 z_0 z_5 + 2 z_1 z_4 + 2 z_1 z_5 + 2 z_2 z_4 + 2 z_2 z_5\\
	   &\quad + 2 z_3 z_4 + 2 z_3 z_5 + 4 z_4 z_5 + 2 z_4^2 + 2 z_5^2 \ne z_0 + z_1 + z_2 + z_3 + 2 z_4 + 2 z_5, \\\\
	   & \left(2 z_0 z_2 - z_1 - z_2 - z_3 - 2 z_4 - 2 z_5 - z_0 + 2 z_0 z_3 + 2 z_1 z_2 + 2 z_0 z_4 + 2 z_1 z_3 + 2 z_0 z_5\right.\\
       &\quad \left. + 2 z_1 z_4 + 2 z_1 z_5 + 2 z_2 z_4 + 2 z_2 z_5 + 2 z_3 z_4 + 2 z_3 z_5 + 4 z_4 z_5 + 2 z_4^2 + 2 z_5^2\right) \times \\
       &\quad \left(z_1 + z_2 + z_4 + z_5 - z_0 z_1 - 2 z_0 z_2 - z_0 z_3 - 3 z_1 z_2 - 2 z_0 z_4 - 2 z_1 z_3 - z_0 z_5 - 4 z_1 z_4\right.\\
       &\quad \left. - z_2 z_3 - 3 z_1 z_5 - 4 z_2 z_4 - 3 z_2 z_5 - 2 z_3 z_4 - z_3 z_5 - 4 z_4 z_5 + 2 z_0 z_2^2 + 2 z_1 z_2^2 + 2 z_1^2 z_2\right.\\
       &\quad \left. + 2 z_0 z_4^2 + 2 z_1^2 z_3 + 4 z_1 z_4^2 + 2 z_1^2 z_4 + 2 z_1 z_5^2 + 4 z_2 z_4^2 + 2 z_1^2 z_5 + 2 z_2^2 z_4 + 2 z_2 z_5^2 \right.\\
       &\quad \left. + 2 z_3 z_4^2 + 2 z_2^2 z_5 + 2 z_4 z_5^2 + 4 z_4^2 z_5 - z_1^2 - z_2^2 - 3 z_4^2 + 2 z_4^3 - z_5^2 + 2 z_0 z_1 z_2\right.\\
       &\quad \left. + 2 z_0 z_1 z_3 + 2 z_0 z_1 z_4 + 2 z_0 z_2 z_3 + 2 z_0 z_1 z_5 + 4 z_0 z_2 z_4 + 2 z_1 z_2 z_3 + 2 z_0 z_2 z_5\right.\\
       &\quad \left. + 2 z_0 z_3 z_4 + 6 z_1 z_2 z_4 + 4 z_1 z_2 z_5 + 4 z_1 z_3 z_4 + 2 z_0 z_4 z_5 + 2 z_1 z_3 z_5 + 2 z_2 z_3 z_4 + 6 z_1 z_4 z_5\right.\\
       &\quad + 2 z_2 z_3 z_5 + 6 z_2 z_4 z_5 + 2 z_3 z_4 z_5) \le 0,\\\\
	   & \left(2 z_0 z_2 - z_1 - z_2 - z_3 - 2 z_4 - 2 z_5 - z_0 + 2 z_0 z_3 + 2 z_1 z_2 + 2 z_0 z_4 + 2 z_1 z_3 + 2 z_0 z_5\right.\\
	   &\quad \left. + 2 z_1 z_4 + 2 z_1 z_5 + 2 z_2 z_4 + 2 z_2 z_5 + 2 z_3 z_4 + 2 z_3 z_5 + 4 z_4 z_5 + 2 z_4^2 + 2 z_5^2\right) \times\\
	   &\quad \left(z_0 + z_3 + z_4 + z_5 - z_0 z_1 - 2 z_0 z_2 - 3 z_0 z_3 - z_1 z_2 - 3 z_0 z_4 - 2 z_1 z_3 - 4 z_0 z_5\right.\\
	   &\quad \left.- z_1 z_4 - z_2 z_3 - 2 z_1 z_5 - z_2 z_4 - 2 z_2 z_5 - 3 z_3 z_4 - 4 z_3 z_5 - 4 z_4 z_5 + 2 z_0^2 z_2\right.\\
	   &\quad \left. + 2 z_0 z_3^2 + 2 z_0^2 z_3 + 2 z_0 z_4^2 + 2 z_1 z_3^2 + 2 z_0^2 z_4 + 4 z_0 z_5^2 + 2 z_0^2 z_5 + 2 z_1 z_5^2 + 2 z_2 z_5^2\right.\\
	   &\quad \left. + 2 z_3 z_4^2 + 2 z_3^2 z_4 + 4 z_3 z_5^2 + 2 z_3^2 z_5 + 4 z_4 z_5^2 + 2 z_4^2 z_5 - z_0^2 - z_3^2 - z_4^2 - 3 z_5^2 + 2 z_5^3\right.\\
	   &\quad \left. + 2 z_0 z_1 z_2 + 2 z_0 z_1 z_3 + 2 z_0 z_1 z_4 + 2 z_0 z_2 z_3 + 2 z_0 z_1 z_5 + 2 z_0 z_2 z_4 + 2 z_1 z_2 z_3 \right.\\
	   &\quad \left. + 4 z_0 z_2 z_5 + 4 z_0 z_3 z_4 + 6 z_0 z_3 z_5 + 2 z_1 z_2 z_5 + 2 z_1 z_3 z_4 + 6 z_0 z_4 z_5 + 4 z_1 z_3 z_5\right.\\
	   &\quad \left. + 2 z_2 z_3 z_4 + 2 z_1 z_4 z_5 + 2 z_2 z_3 z_5 + 2 z_2 z_4 z_5 + 6 z_3 z_4 z_5\right) \le 0,\\\\
	   \text{and} \quad& 0 \le z_0, z_1, z_2, z_3, z_4, z_5\le 1.
	\end{align*}
The above analytic solution to \eqref{eq:z:first}--\eqref{eq:z:last} is obtained by symbolic math programming in MATLAB.

\bibliography{17-492_initials_only}

\end{document}